\documentclass{jmmrc}
\begin{document}

\title[]{Generating Modern Persian Carpet Map by Style-transfer}

	\author{Dorsa Rahmatian}
\address{	
	\newline Orcid number: 
	\newline Department of Computer Engineering
	\newline Shahid Bahonar University of Kerman
	\newline Kerman, Iran}
\email{dorsa.rahmatian@eng.uk.ac.ir}

	\author{Monireh Moshavash}
\address{	
	\newline Orcid number: 
	\newline Department of Computer Engineering
	\newline Shahid Bahonar University of Kerman
	\newline Kerman, Iran}
\email{monireh.moshavash@gmail.com}

	\author{Mahdi Eftekhari*}
	\address{
			\newline Orcid number: 
			\newline Department of Computer Engineering
		\newline Shahid Bahonar University of Kerman
		\newline Kerman, Iran}
	\email{m.eftekhari@uk.ac.ir}

	\author{Kamran Hoseinkhani}
\address{
	\newline Orcid number: 
	\newline Department of Carpet, Saba Faculty of Art and Architecture
	\newline Shahid Bahonar University of Kerman
	\newline Kerman, Iran}
\email{k.hoseinkhani@uk.ac.ir}

\thanks{\hspace*{-0.60cm} {\footnotesize\noindent m.eftekhari@uk.ac.ir, 
ORCID: 0000-0002-0381-8225,\\  
}
{\footnotesize DOI: 10.22103/jmmr.2023.20648.1373 
\hspace{3cm}\copyright{ the Author(s)}\\
Publisher: Shahid Bahonar University of Kerman\\
How to cite: D. Rahmatian, M. Moshavash, M. Eftekhari and K. Hoseinkhani, {\it Generating Modern Persian Carpet Map by Style-transfer}, J. Mahani Math. Res. 2024; 13(1): 23-44.
		}
	}

\fancyhead[CO]{\tiny{Generating Modern Persian Carpet by Style-transfer -- {JMMR Vol. 13, No. 1 (2024)}}}
\fancyhead[CE]{\tiny{ D. Rahmatian, M. Moshavash, M. Eftekhari and K. Hoseinkhani}}

\maketitle

\begin{center}
	{\tiny Article type: Research Article \\(Received: 04 December 2022, Received in revised form 23 February 2023)}\\
	{\tiny (Accepted: 10 April 2023, Published Online: 10 April 2023 )}
\end{center}

\begin{abstract}
Today, the great performance of Deep Neural Networks(DNN) has been proven in various fields. One of its most attractive applications is to produce artistic designs. A carpet that is known as a piece of art is one of the most important items in a house, which has many enthusiasts all over the world. The first stage of producing a carpet is to prepare its map, which is a difficult, time-consuming, and expensive task.
In this research work, our purpose is to use DNN for generating a Modern Persian Carpet Map.  To reach this aim,  three different DNN style transfer methods are proposed and compared against each other. In the proposed methods, the Style-Swap method is utilized to create the initial carpet map, and in the following, to generate more diverse designs, methods Clip-Styler, Gatys, and Style-Swap are used separately. In addition, some methods are examined and introduced for coloring the produced carpet maps. The designed maps are evaluated via the results of filled questionnaires where the outcomes of user evaluations confirm the popularity of generated carpet maps. Eventually, for the first time, intelligent methods are used in producing carpet maps, and it reduces human intervention. The proposed methods can successfully produce diverse carpet designs, and at a higher speed than traditional ways.\\
\end{abstract}

\section{Introduction}

The strong performance of Deep Neural Networks has been demonstrated in various fields of AI. 
Neural Style Transfer(NST) is a popular field and attracts attention in recent years.
Style transfer means transferring the style and texture of one image(style image) to another image(content image). As the output image is a content image that has the style of the style image. NST employs convolutional neural networks(CNN) as a feature extractor to provide features of style image and content image \cite{gatys2016image}. Before neural networks, texture transfer methods were used for transferring style to the content images e.g. they employed MRF \cite{ashikhmin2003fast} in finding appropriate pixels \cite{zhao2020survey}. 
Carpet is one of the important parts of house design. Iran is one of the countries that has a world reputation for producing carpet maps. Designing carpet maps in the traditional manner is done in three steps that are as follows: 1) Drawing the initial plan, 2) Dotting the carpet, and 3) Colorizing the final map.  The first step is performed manually by an expert person, and the other steps are done through photo-shop software. Producing a carpet map in this way is a time-consuming and expensive process. Also, these designs do not have a lot of variety.  

As mentioned above, designing a carpet with current methods is a difficult task. To improve this issue, for the first time, we applied the style transfer methods for generating a modern Persian carpet map. In this work, we have applied some style transfer methods as tools and have introduced three algorithms for producing  carpet maps. Our contributions are as follows: 
\begin{itemize}
	\item For the first algorithm, two methods Style-Swap and Clip-Styler are used. By putting them together in a special way, a novel method for generating carpet maps is introduced.
	
	\item In the second algorithm, two methods (Style-Swap, and Gatys) are applied for generating a new Persian carpet design. 
	
	\item Finally, one method is used twice (Style-Swap, and Style-Swap) for making a new Persian carpet design.
	
	\item One demanding step of creating the carpet map is its colorization. In this research, fast and efficient methods are proposed for colorizing carpet maps.
	
\end{itemize}
	
\section{Related work}
Texture transfer is a long-term research field that has attracted the attention of computer and art researchers since years ago. Recently deep neural networks have shown superior performance in various fields of artificial intelligence. 
The neural style transfer technique has been introduced by Gatys et al. \cite{gatys2016image} and it has employed deep neural networks in texture transfer tasks. They used convolutional neural networks to generate stylized photos. Gatys et al. \cite{gatys2016image} discovered that the content and style of an image can be obtained from features extracted from convolutional neural network layers. After that many neural style transfer works have been done. generally, there are two categories of neural style transfer methods which transfer style locally \cite{sheng2018avatar,gu2018arbitrary,li2016combining} or globally \cite{li2017universal,li2019learning,huang2017arbitrary}. Methods that transfer style globally do not consider the local patterns and their locations but they transfer the general style of the style image well. In contrast, the methods which transfer style locally consider the details of style image and transfer them.
Huang et al. \cite{huang2017arbitrary} proposed a global style transfer method that used domain adaptation. They realized that aligning the mean and variance of content features with those of style features produced acceptable stylized results.
Zhang et al. \cite{zhang2022domain} presented a method that, unlike some previous methods(\cite{gatys2016image}), instead of using second-order statistics(like gram matrix of features), uses the image features for learning style representation.
Style-Swap \cite{chen2016fast} is an example of the local style transfer method. It divides the style and content features into equal-size patches and finds the most appropriate style patch for each content patch using a cross-correlation measure.
Recently some methods have been proposed to take the advantage of each category and tackle their issues, e.g. Zhang et al. \cite{zhang2019multimodal} used the clustering method on style features and clustered similar style patterns, then finds the semantic matching using graph-cut formulation, and transfers style of each cluster to appropriate content features. The advantage of using the clustering operation is that it groups similar style features regardless of their spatial locations. Afifi et al.\cite{afifi2021cams} have presented the "Color-Aware Multi-Style Transfer (Cams)"  method that considers the correlation between styles and colors which means that in their method the texture of color in the style image has been transferred to the nearest color of the content image.

Kwon et al. \cite{kwon2022clipstyler} proposed a different framework that takes a text description of style instead of an image and transfers it to the content image. Also, Liu et al. \cite{liu2022name} used the CLIP model \cite{radford2021learning} and introduced another text-driven image style transfer method that could transfer the specific artistic characters to the content image.
Karras et al. \cite{karras2019style} introduced style-gan which employs the style transfer task in generative adversarial networks to generate face images.
Zhu et al. \cite{zhu2017unpaired} proposed an image-to-image translation without the need for a large paired dataset by only using a cycle consistency loss.

In this work, we choose some style transfer methods for generating novel carpet maps and to the best of our knowledge, no similar work has been done in this field before.

\section{Preliminaries} \label{Preliminaries}
\subsection{Gatys}
\begin{figure*}
	\centering
	\includegraphics[width=\linewidth]{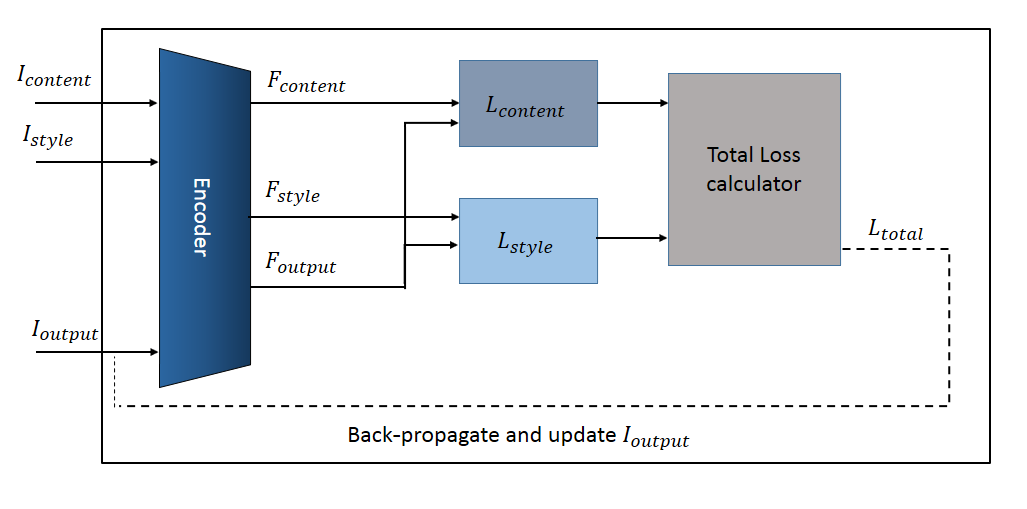}
	\caption{Overall schematics of Gatys et al. method}
	\label{fig:gatys}
\end{figure*}

Gatys et al. \cite{gatys2016image} introduced neural style transfer for the first time. Due to the impressive performance(operation) of deep convolutional neural networks in feature extraction tasks, Gatys et al. \cite{gatys2016image} employed the VGG-16 network to get the style and content of an image. They proposed an image-optimization-based model which uses a pre-trained VGG-16 network to obtain the style and content of an image using extracted features. Image-optimization-based models are a group of style transfer models that update an image that they have received as input($I_{output}$) during several epochs according to the defined loss function using the back-propagation operation.
Gatys et al. discovered that the features of deep layers of a CNN represent the main content of the image, also the correlation of feature channels(called Gram matrix) obtained from a CNN indicates the style of the image. Based on the above-mentioned exploration, the total loss consists of two style and content parts as follows:
\begin{equation}
	\mathcal{L}_{\text {content }}=\frac{1}{2} \sum\left(F_{content}^{l}-F_{output}^{l}\right)^{2}
	\label{eq:content-loss}
\end{equation}
\begin{equation}
	\mathcal{L}_{\text {style }}=\frac{1}{4 C_{l}^{2}\left(H_{l} W_{l}\right)^{2}} \sum\left(G_{F_{style}}^{l}-G_{F_{output}}^{l}\right)^{2}
	\label{eq:style-loss}
\end{equation}
\begin{equation}
	\mathcal{L}_{\text {total }}=\mathcal{L}_{\text {content }}+\lambda \mathcal{L}_{\text {style }}
	\label{eq:total-loss}
\end{equation}
where $F_{x}^{l} \in \mathcal{R}^{C_{l} \times H_{l} \times W_{l}}$ and $G_{F_{x}}^{l}$ indicate the features and gram matrix of features of image x obtained from layer $l$ of the VGG encoder respectively. For $\mathcal{L}_{\text {content }}$ feature extracted from layers Relu-$x$-1 , $x \in [4,5]$ and for $\mathcal{L}_{\text {style }}$ feature extracted from layer Relu-$x$-1 , $x \in [1,5]$ is employed.
After computing the total loss, the stylized($I_{output}$) image will be optimized using a back-propagation algorithm and returned to the network as input(Fig. \ref{fig:gatys}).

\subsection{Cams}
Afifi et al.\cite{afifi2021cams} introduced another image-optimization-based method. They proposed an approach that considers the correlation between styles and colors which means that in their method the texture of one color in the style image has been transferred to the nearest color of the content image. This method produces more pleasant results than Gatys when the style image has multiple style patterns. Because Gatys presents a mixture of all styles all over the output image while the Cams method transfers a specific texture and style to a certain color(region) in the content image.
The most significant difference between Gatys and Cams methods is in their style loss since the Cams method generates weighted gram matrices based on color similarities while the Gatys method generates a global gram matrix. In the following, the steps of producing a weighted Gram matrix have been explained.

\begin{figure*}
	\centering
	\includegraphics[width=\linewidth]{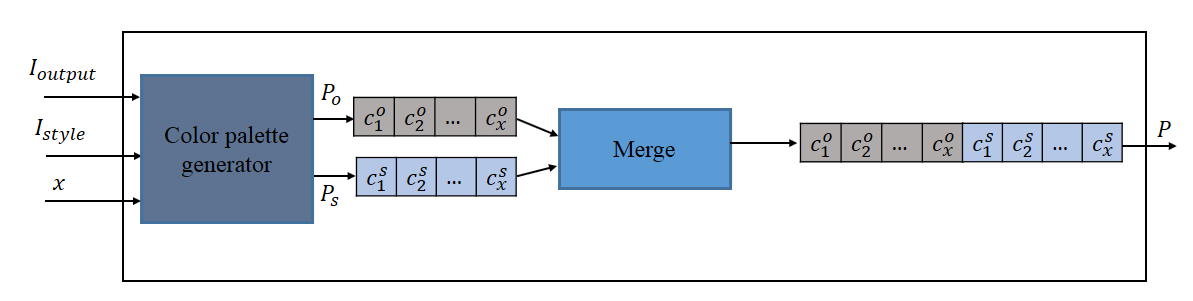}
	\caption{Producing color palette $P$ from $I_{output}$ and $I_{style}$.}
	\label{fig:cams1}
\end{figure*}

The Cmas method initializes the output image($I_{output}$) by content image($I_{content}$) and utilizes the algorithm proposed in \cite{chang2015palette} to extract the color palette with size $x$ from the output image and style image($I_{style}$) separately and then merges two palettes to obtain one single palette $P$ (Fig. \ref{fig:cams1}).

Then two weighting mask sets($M_{output},M_{style}$) are created for the output image and style image using all colors in the palette $P$. The weighting masks are used to add weights to the feature maps of output and style images. The main step of Cams is computing a gram matrix using weighted features for each color $t \in P$. 
Then the style loss is calculated using the gram matrices obtained from each color separately according to the following equation

\begin{equation}
	\mathcal{L}_{style}=
	\sum_{{l}}\sum_{\mathrm{t}}\left\|G_{F_{style}^{\mathrm{t}}}^{l}-G_{F_{output}^{\mathrm{t}}}^{l}\right\|_2^2 .
\end{equation}

where $G_{F_{I}^{\mathrm{t}}}$ indicates the gram matrix computed from weighted features of the image $I$ using color $t \in P$. Similar to the Gatys method to compute the total style loss features extracted from layers Relu-$x$-1 , $x \in [1,5]$ have been employed.

For the content loss, they use content loss introduced in Gatys method(Eq. \ref{eq:content-loss}. The final total loss will be similar to Eq. \ref{eq:total-loss}.

\begin{figure*}
	\centering
	\includegraphics[width=\linewidth]{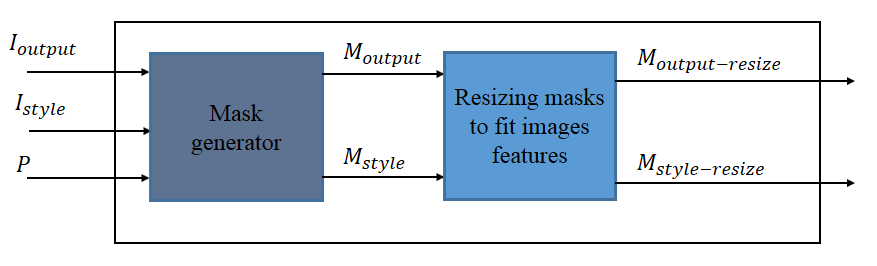}
	\caption{Generating two weighting mask sets $M_{output}$ and $M_{style}$ for $I_{output}$ and $I_{style}$ respectively using all color $t \in P$ where 
		$M_{\text {I}}=\left\{mask_{\text {I}}^t\right\} \quad t \in \text { palette }$ for image $I$. Then they will be resized($M_{output-resize}$ and $M_{style-resize}$) to be the same size as extracted features of the encoder to be employed in producing weighted feature maps. }
	\label{fig:cams2}
\end{figure*}
\begin{figure*}
	\centering
	\includegraphics[width=\linewidth]{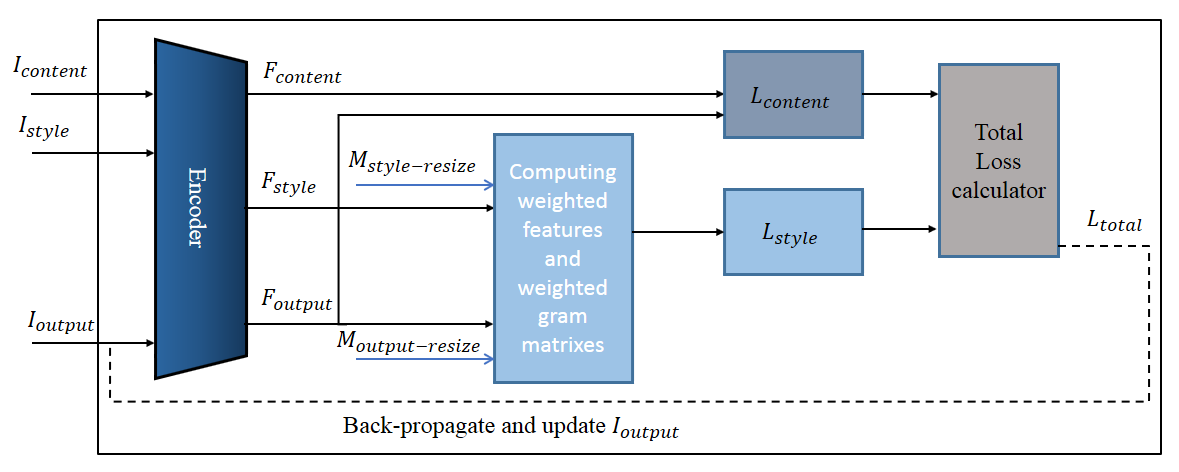}
	\caption{Overall schematics of Cams method}
	\label{fig:cams3}
\end{figure*}

\subsection{Style-Swap}
\begin{figure*}
	\centering
	\includegraphics[width=\linewidth]{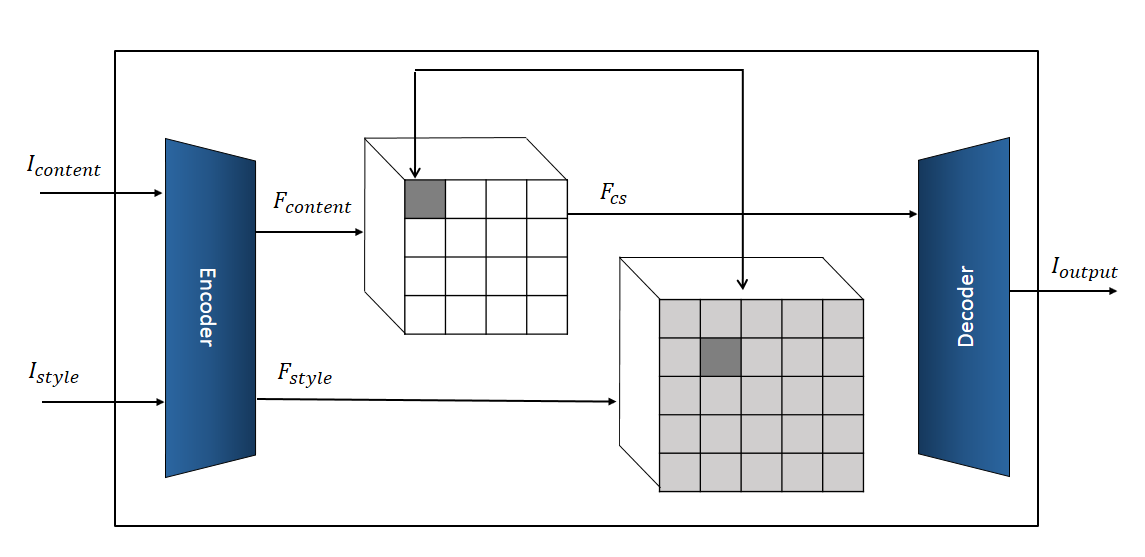}
	\caption{Overall schematics of Style-Swap method}
	\label{fig:style_swap}
\end{figure*}

Chen et al. \cite{chen2016fast} proposed a feed-forward method composed of a pre-trained auto-encoder containing a VGG-19 as an encoder and a decoder that has the symmetric structure of the VGG encoder. After feeding a content image and a style image to the encoder part, the features of each image are extracted from a certain layer(conv4-1). Features of content and style images are divided into patches of equal size. After patching features, to find the best style patch for a content patch, the cross-correlation measure is implemented by a convolution layer. Then each patch of the content features is replaced by the best-matching style patch. The final result features are fed to the decoder to create the stylized output image(Fig. \ref{fig:style_swap}). 

\subsection{Clip-Styler}
The main difference between the method proposed in Kwon et al. \cite{kwon2022clipstyler} paper and most other style transfer methods is it takes a text as the style instead of an image. Radford et al. \cite{radford2021learning} proposed the CLIP model that transfers the information of the text to the visual space. Recently, Kwon et al. \cite{kwon2022clipstyler} have employed an embedding model of the CLIP in style transfer tasks. Also, they use a trainable lightweight CNN to generate the stylized image. The directional loss function Eq.\ref{eq:clip_loss3} introduced by Style\_gan \cite{gal2021stylegan} is utilized to align the direction of the output style text with the target style text and also output content with the target content image in the CLIP latent space.

\begin{equation}
	\Delta T=E_{T}\left(T_{style}\right)-E_{T}\left(T_{content}\right)
	\label{eq:clip_loss1}
\end{equation}
\begin{equation}
	\Delta I=E_{I}\left(I_{output}\right)-E_{I}\left(I_{content}\right)
	\label{eq:clip_loss2}
\end{equation}

\begin{equation}
	L_{CLIP}=1-\frac{\Delta I \cdot \Delta T}{|\Delta I \| \Delta T|}
	\label{eq:clip_loss3}
\end{equation}
where $E_{I}$ and $E_{T}$ indicate the image and text encoders of CLIP, respectively; and $T_{style}$ and $T_{content}$ show the semantic text of the style target and the input content respectively. If the content image does not have any style text, $T_{content}$ will be set as "Photo". $I_{content}$ and $I_{output}$ are the input content and the stylized output images.

\begin{figure*}
	\centering
	\includegraphics[width=\linewidth]{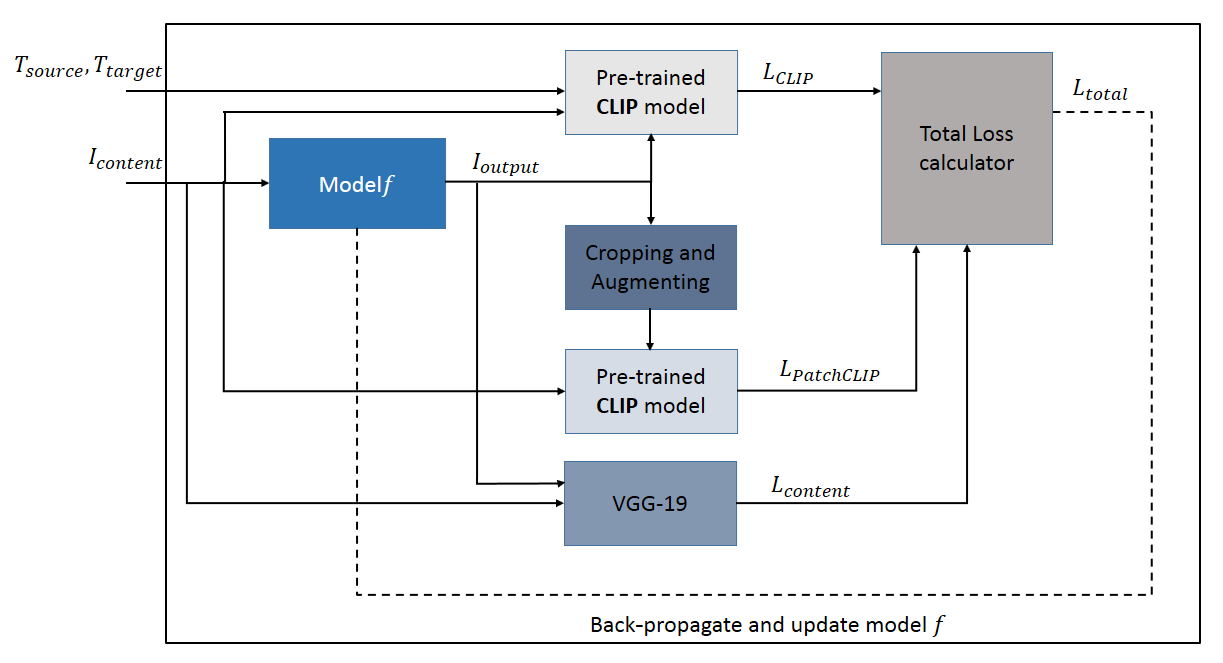}
	\caption{Overall schematics of Clip-Styler method}
	\label{fig:CLIP_styler}
\end{figure*}

Also, they proposed a PatchCLIP loss to transfer the local semantic texture of the $T_{style}$ to the content image. To do this, the stylized image is cropped to some equal-sized patches, and a perspective augmentation is applied on each one, then the CLIP loss will be computed for each augmented patch. To avoid over-stylization of some specific patches, they use a regularization term that will neutralize the losses of those patches. To increase the importance of content preservation, the content loss function introduced by Gatys et al. \cite{gatys2016image} is also employed(described in Eq. \ref{eq:content-loss}). Finally, the total loss is computed from the summation of the above-mentioned components and
the lightweight CNN $f$ will be optimized by back-propagating the total loss after each iteration. After some iteration output of the CNN model, $f$ will be demonstrated as the final stylized result. An overall preview of the Clip-Styler method is shown in Fig. \ref{fig:CLIP_styler}. 

\section{Proposed method}\label{sec:proposedmethod}
In this paper style transfer methods are employed as tools for generating modern carpet maps. We proposed three different algorithms based on the style transfer methods which are mentioned in Sec. \ref{Preliminaries}. In addition, some suitable style transfer methods have been introduced for colorizing generated carpet maps. In the following sections, the proposed algorithms are expressed in detail.

\subsection{Generating Carpet Map}\label{sec:generating}
The main idea of this paper is to put some style transfer methods consecutively, to generate modern carpet maps.
 
Preferring to generate more different carpet maps from the initial one, we use two style transfer methods in a row. 
In all the proposed methods, we used the Persian carpet map as the initial content input, then new style inputs were employed to generate modern carpet maps.

In the first step, we obtain the new carpet map's initial format with a specific style by the style-Swap method.
In the first stage, the Style-Swap method is used which is a local style transfer method to preserve the general template of the Persian carpet.
The patch size and stride of this method have a fundamental role in the generated carpet maps. The higher the patch size, the lower the details of the initial carpet map in the resulting image, and vice versa. 
Then, in the second step, to add details and make more differences in the new map, we use another style transfer method with other style images. According to our research, in this paper, we suggest using three methods Style-Swap, Gatys, and Clip-Styler, in the second step because they can produce modern, various, and new carpet designs.
\begin{figure*}
	\centering
	\includegraphics[width=\linewidth]{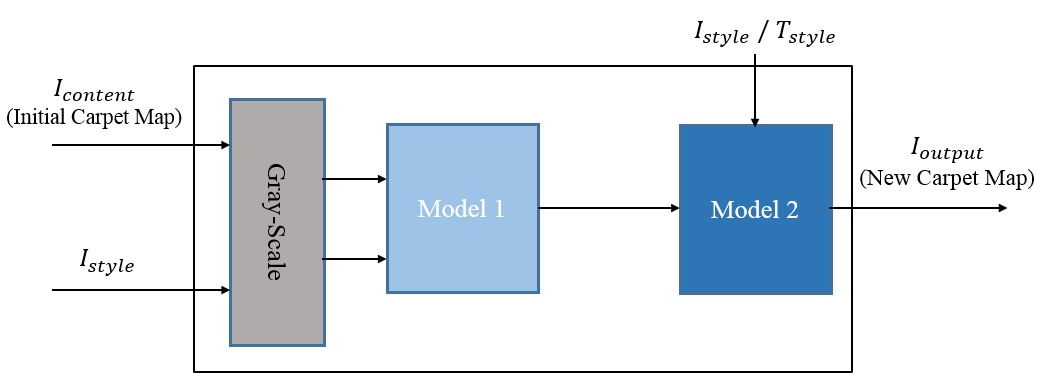}
	\caption{Flowchart of generating a carpet map: Model1 is the Style-Swap method for all proposed algorithms while Model2 is  Clip-Styler, Style-Swap, or Gatys in each one of the proposed algorithms.}
	\label{fig:4}
\end{figure*}

\subsection{Colorization}\label{sec:colorization}
Finally, to colorize the produced grayscale carpet design, style transfer methods are used with the colored style images as their inputs. We found out with trial and error that Cams, Gatys, and Clip-Styler methods are appropriate choices for colorization.

\SetKwComment{Comment}{/* }{ */}
\begin{algorithm}[hbt!]
\caption{Algorithm of the proposed methods}\label{alg:two}
\SetKwInput{KwInput}{Input}                
\SetKwInput{KwOutput}{Output}              
\DontPrintSemicolon
\KwInput{$I_{c}$,$I_{s1}$,$I_{s2}$,$I_{s3}$,$SecondMethod$,$ColoringMethod$}
\Comment*[r]{$I_{c}$=initial carpet image}
\Comment*[r]{$I_{s1}$=first style pattern}
\Comment*[r]{$I_{s2}$=second style pattern or style text}
\Comment*[r]{$I_{s3}$=input colored image}
\Comment*[r]{$SecondMethod$=second style transfer model}
\Comment*[r]{$ColoringMethod$=colorization style transfer model}
\KwOutput{$I_{final}$}
\SetKwFunction{FMain}{Generating}
	
	\SetKwProg{Fn}{Function}{:}{\KwRet}
	\Fn{\FMain}{
$I_{o1} \gets StyleSwap(I_{c},I_{s1})$ \;
\If{$SecondMethod$ = StyleSwap }{
      $I_{o2} \gets StyleSwap(I_{o1},I_{s2})$ \;
    }
\If{$SecondMethod$ = ClipStyler }{
      $I_{o2} \gets ClipStyler(I_{o1},I_{s2})$ \;
    }
\If{$SecondMethod$ = Gatys }{
      $I_{o2} \gets Gatys(I_{o1},I_{s2})$ \;
    }
    }
\SetKwFunction{FMain}{Colorizing}
	
	\SetKwProg{Fn}{Function}{:}{\KwRet}
	\Fn{\FMain}{
\If{$ColoringMethod$ = ClipStyler }{
      $I_{final} \gets ClipStyler(I_{o2},I_{s3})$ \;
    }
\If{$ColoringMethod$ = Cams }{
      $I_{final} \gets Cams(I_{o2},I_{s3})$ \;
    }
\If{$ColoringMethod$ = Gatys }{
      $I_{final} \gets Gatys(I_{o2},I_{s3})$ \;
    }
    }
\SetKwFunction{FMain}{main}
	
	\SetKwProg{Fn}{Function}{:}{\KwRet}
	\Fn{\FMain}{
 $ I_{final} \gets $ Colorizing(Generating($I_{c},I_{s1},I_{s2},SecondMethod$),$I_{s3},ColoringMethod$)
 }
\end{algorithm}

\section{Experiments and results}\label{exp}
\subsection{Experiment setup}
In this work pre-trained models of four algorithms including \cite{chen2016fast,kwon2022clipstyler,afifi2021cams,gatys2016image} are used. Chen et al. \cite{chen2016fast}, Afifi et al. \cite{afifi2021cams}, and Gatys et al. \cite{gatys2016image} used a pre-trained VGG-16 network in their models to extract features of images. Chen et al. \cite{chen2016fast} also used a decoder that has the symmetric layers of a VGG encoder and is pre-trained on the image reconstruction tasks. This decoder is used to transform the embedded features into visual images.
As mentioned in Sec 4.2, Clip-Styler\cite{kwon2022clipstyler} is a style transfer method that gets a text style instead of an image style. Kwon et al. \cite{kwon2022clipstyler} have employed the CLIP model \cite{radford2021learning} which is pre-trained on text-image datasets (\cite{lin2014microsoft,krishna2017visual,thomee2016yfcc100m}) and also a simple CNN which is supposed to learn during the style transferring an image. As said before,\cite{kwon2022clipstyler,afifi2021cams,gatys2016image} have iterative structures in which we set the number of iterations to 500, 300, and 10 respectively. Most of the results in this work have been obtained by setting patch size and stride in Style-Swap \cite{chen2016fast} method 5 and 3 respectively and palette size in Cams \cite{afifi2021cams} method 5.
In the Generating Carpet Map part of our model, the traditional Persian carpet maps are utilized as content images, and Persian Eslimi maps and fractal patterns are used as style images.
Moreover, in the Colorization part, we have employed the obtained new carpet maps as content images and colored style images(e.g. paintings by famous artists) as style images.
\subsection{Results}

\begin{figure*}[h!]
	\centering
	\includegraphics[width=5cm]{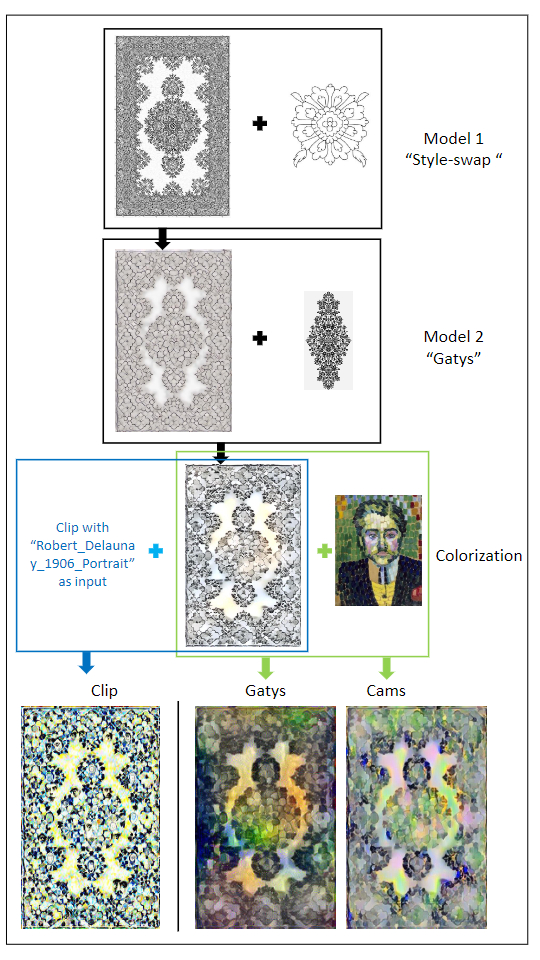}
	\includegraphics[width=5cm]{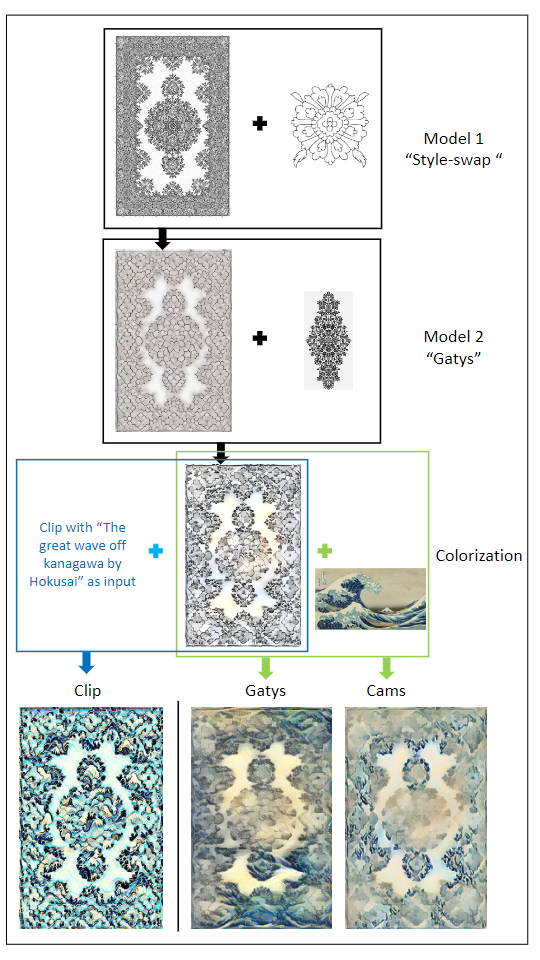}
	\includegraphics[width=5cm]{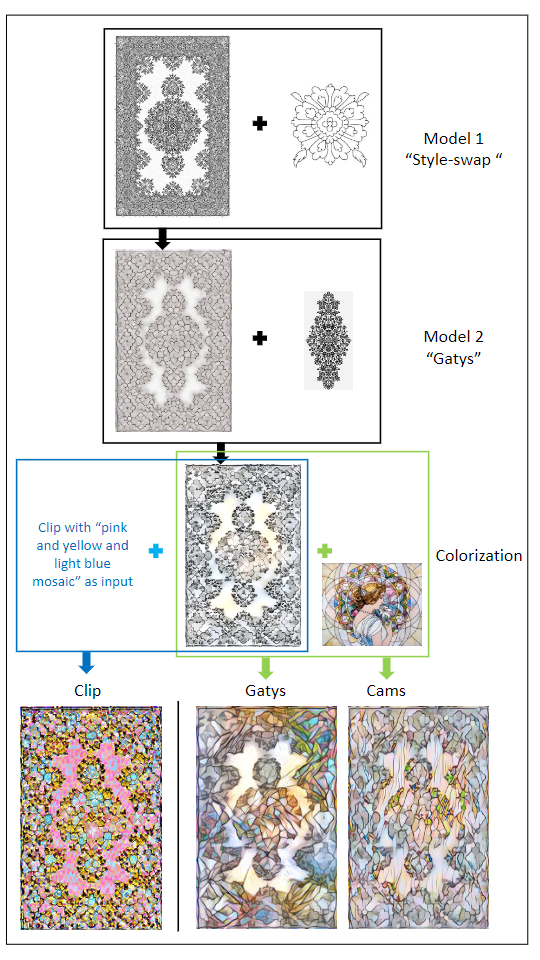}
	\includegraphics[width=5cm]{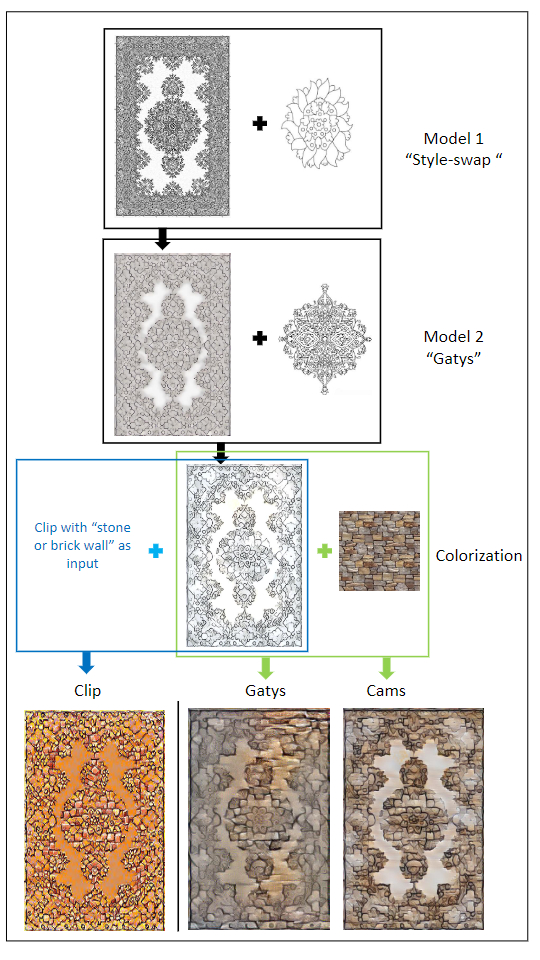}
	\caption{Results of generating and coloring new carpet maps when the first method is "Style-Swap" and the second is "Gatys".}
\label{fig:swap-gatys}
\end{figure*}

\begin{figure*}[h!]
\centering
\includegraphics[width=5cm]{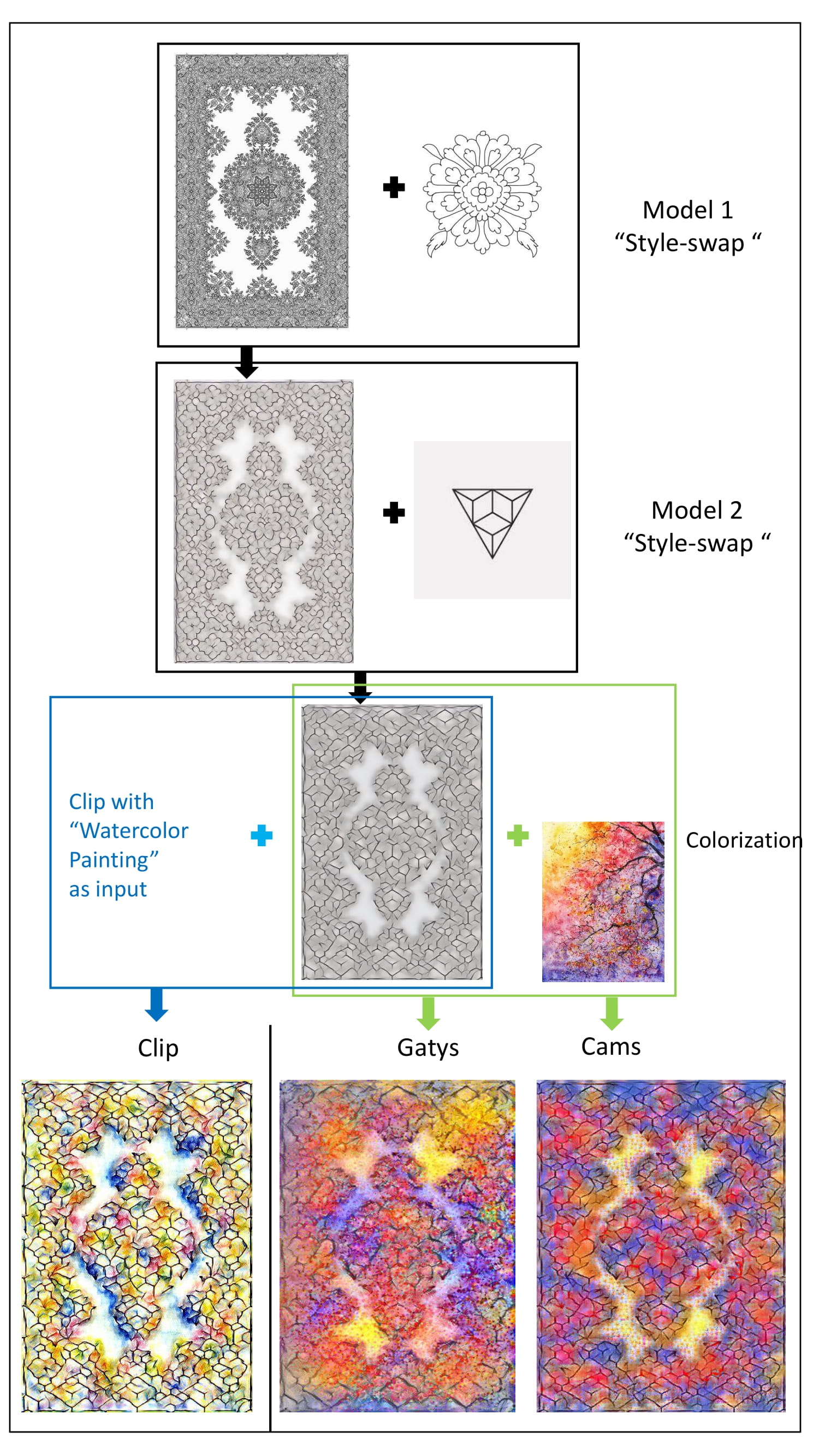}
\includegraphics[width=5cm]{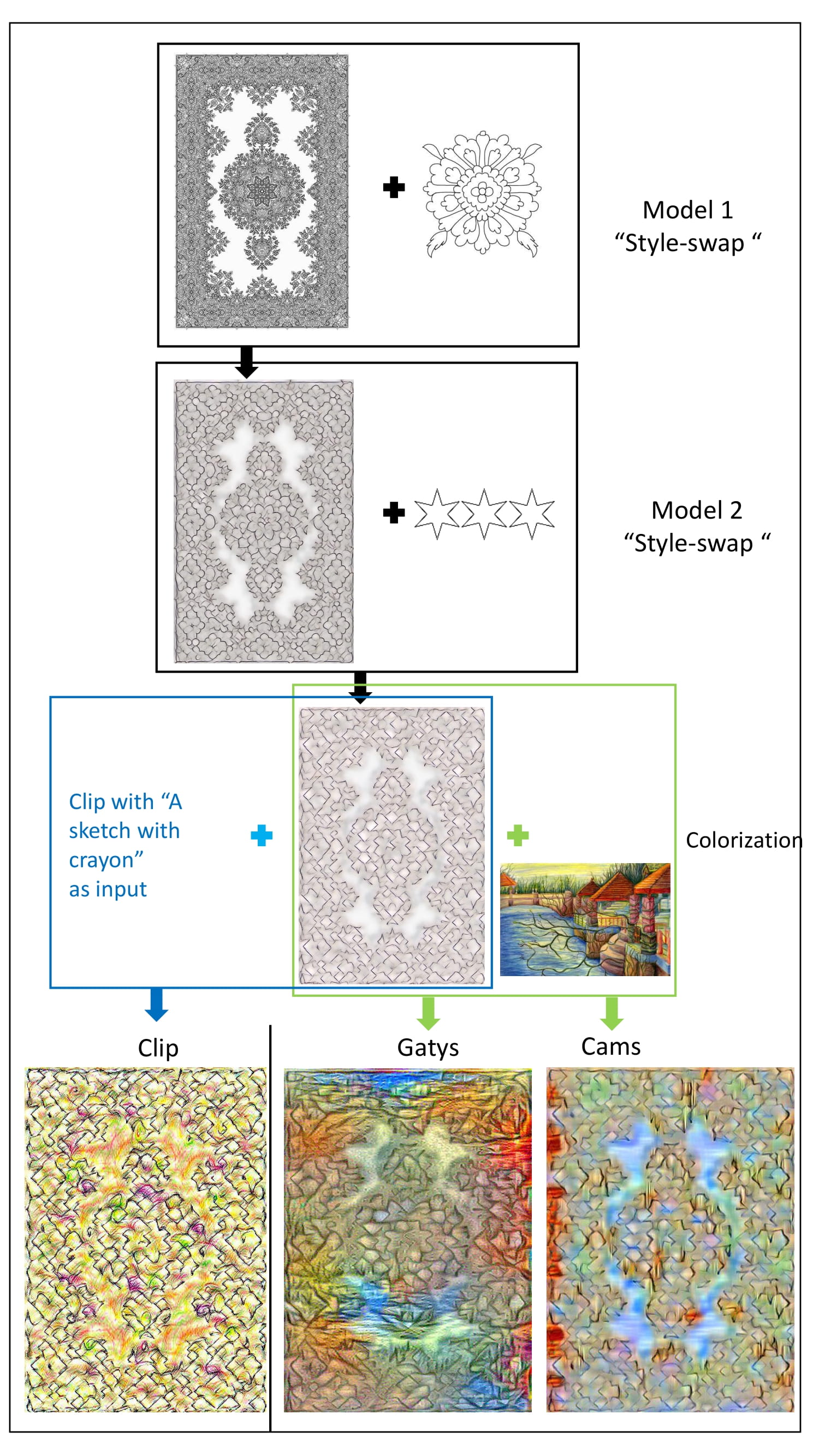}
\includegraphics[width=5cm]{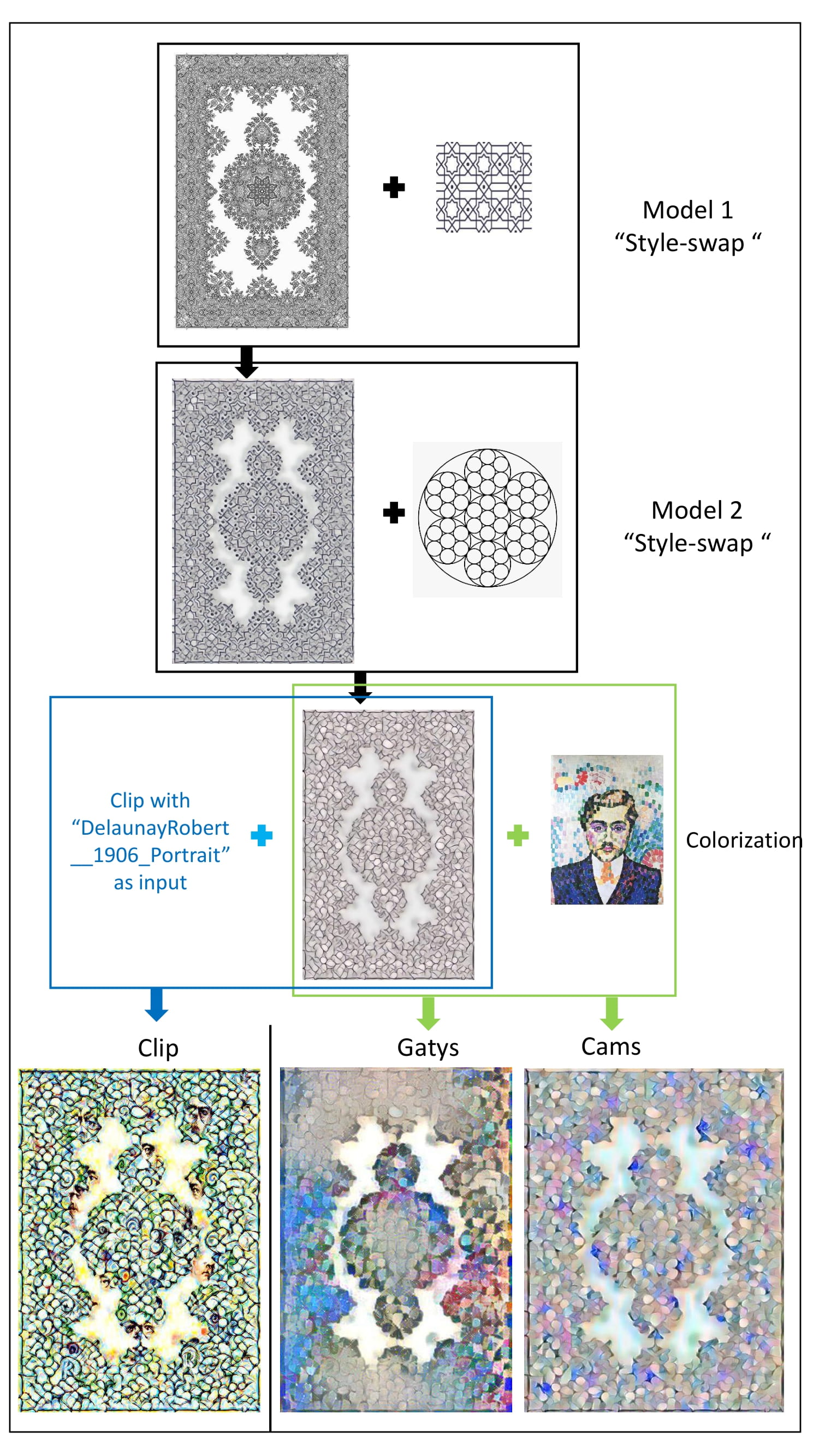}
\includegraphics[width=5cm]{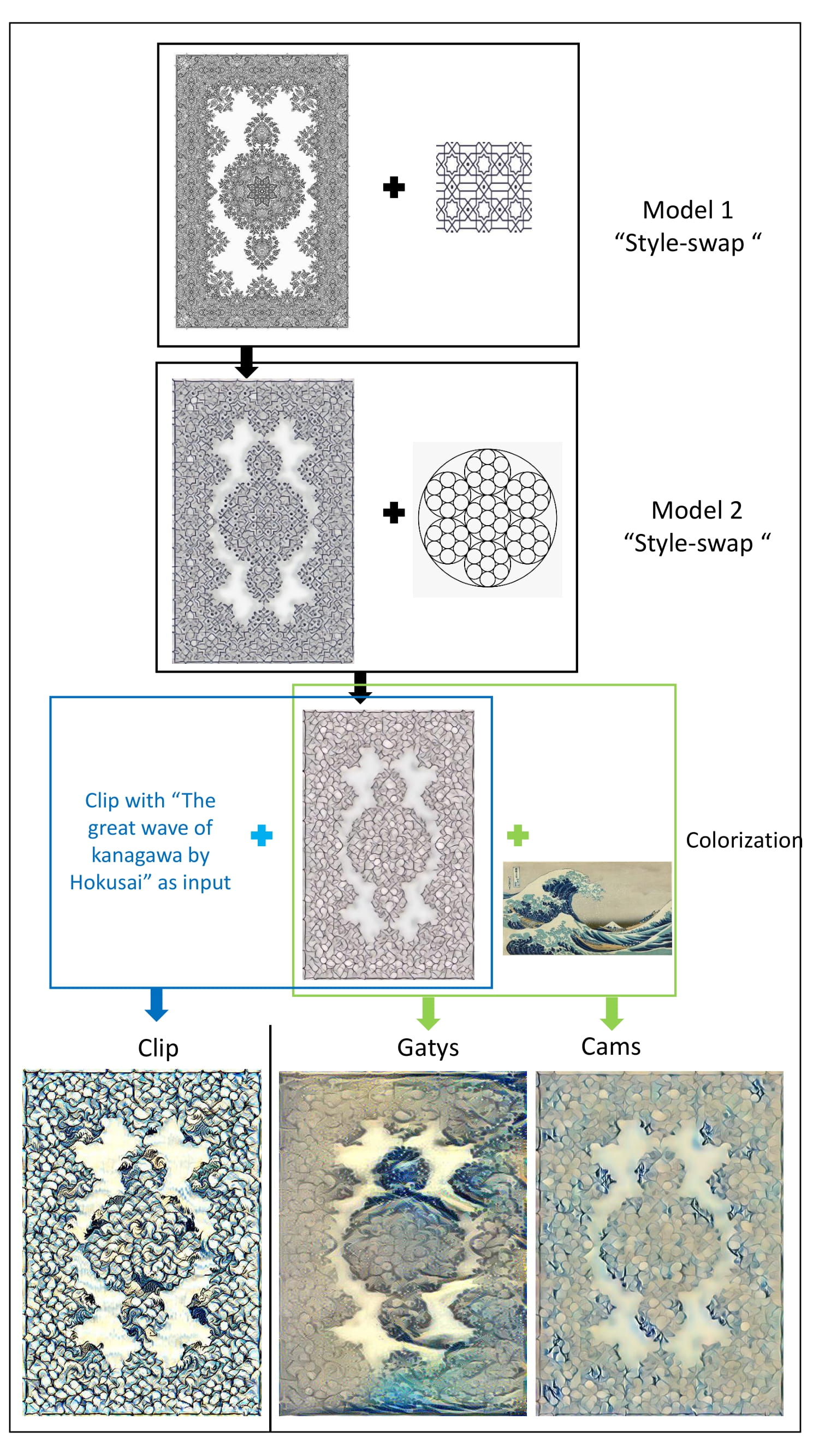}
\caption{Results of generating and coloring new carpet maps when the both first and second methods are "Style-Swap".}
\label{fig:swap-swap}
\end{figure*}

\begin{figure*}[h!]
\centering
\includegraphics[width=5cm]{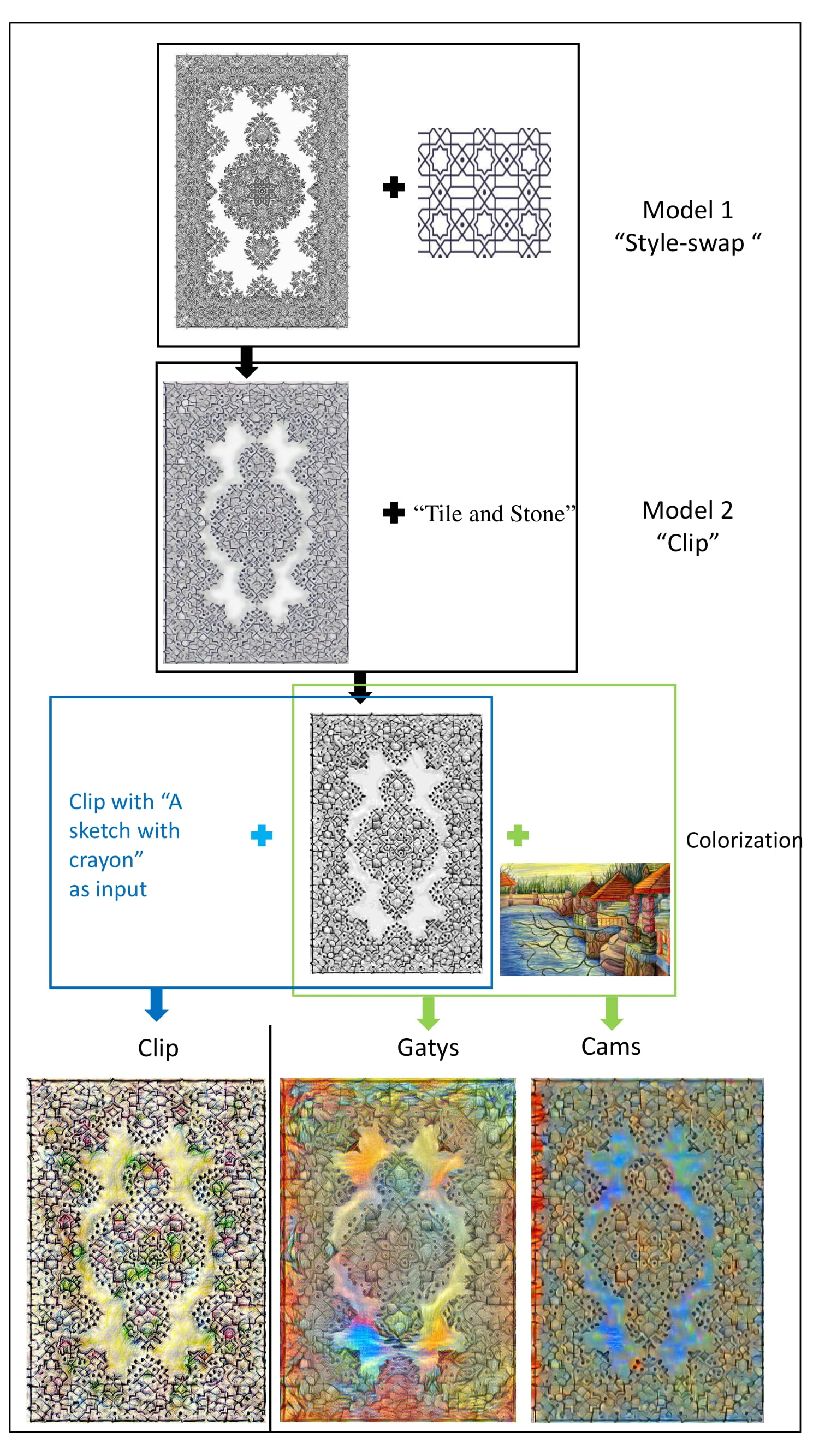}
\includegraphics[width=5cm]{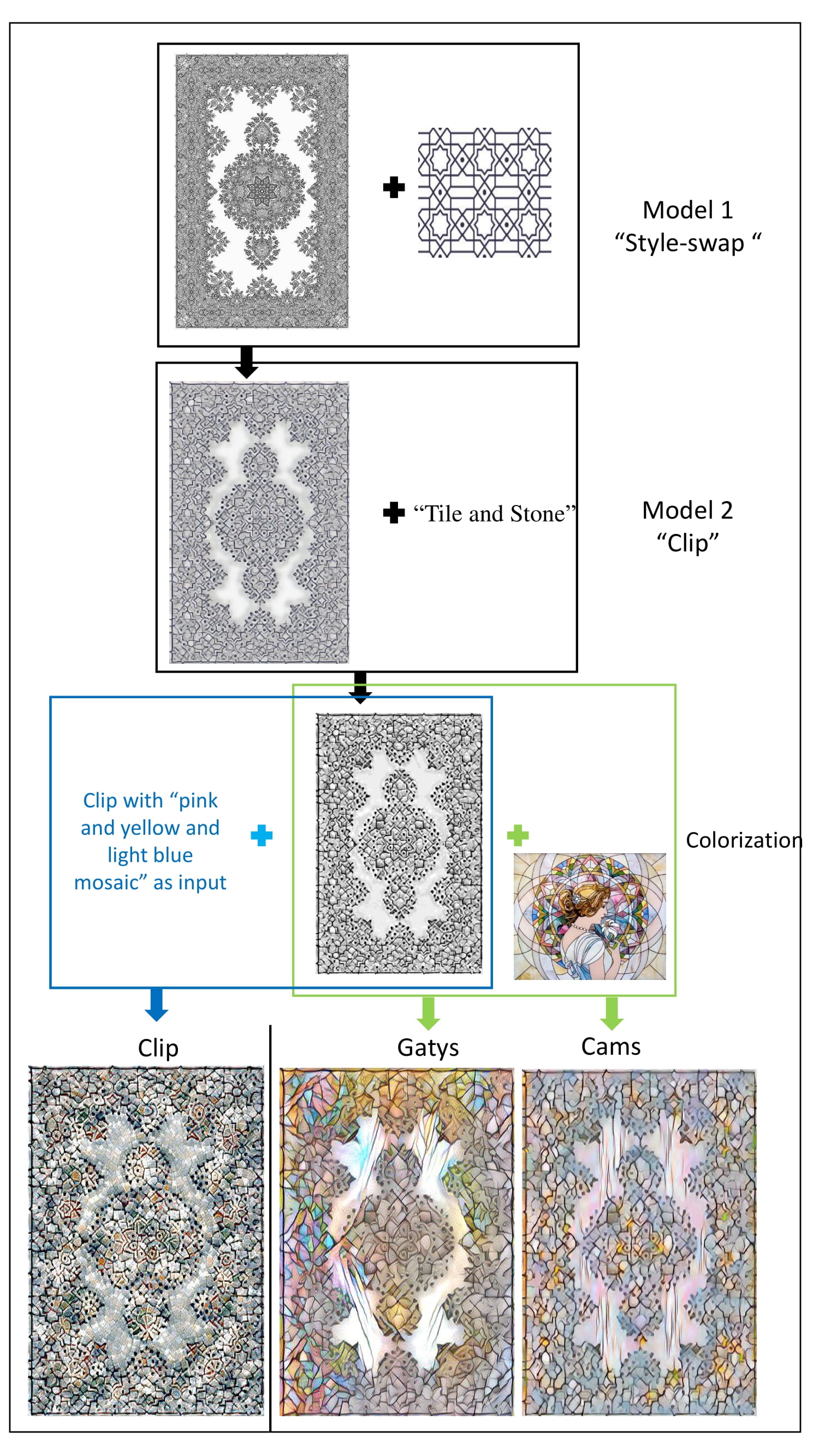}
\includegraphics[width=5cm]{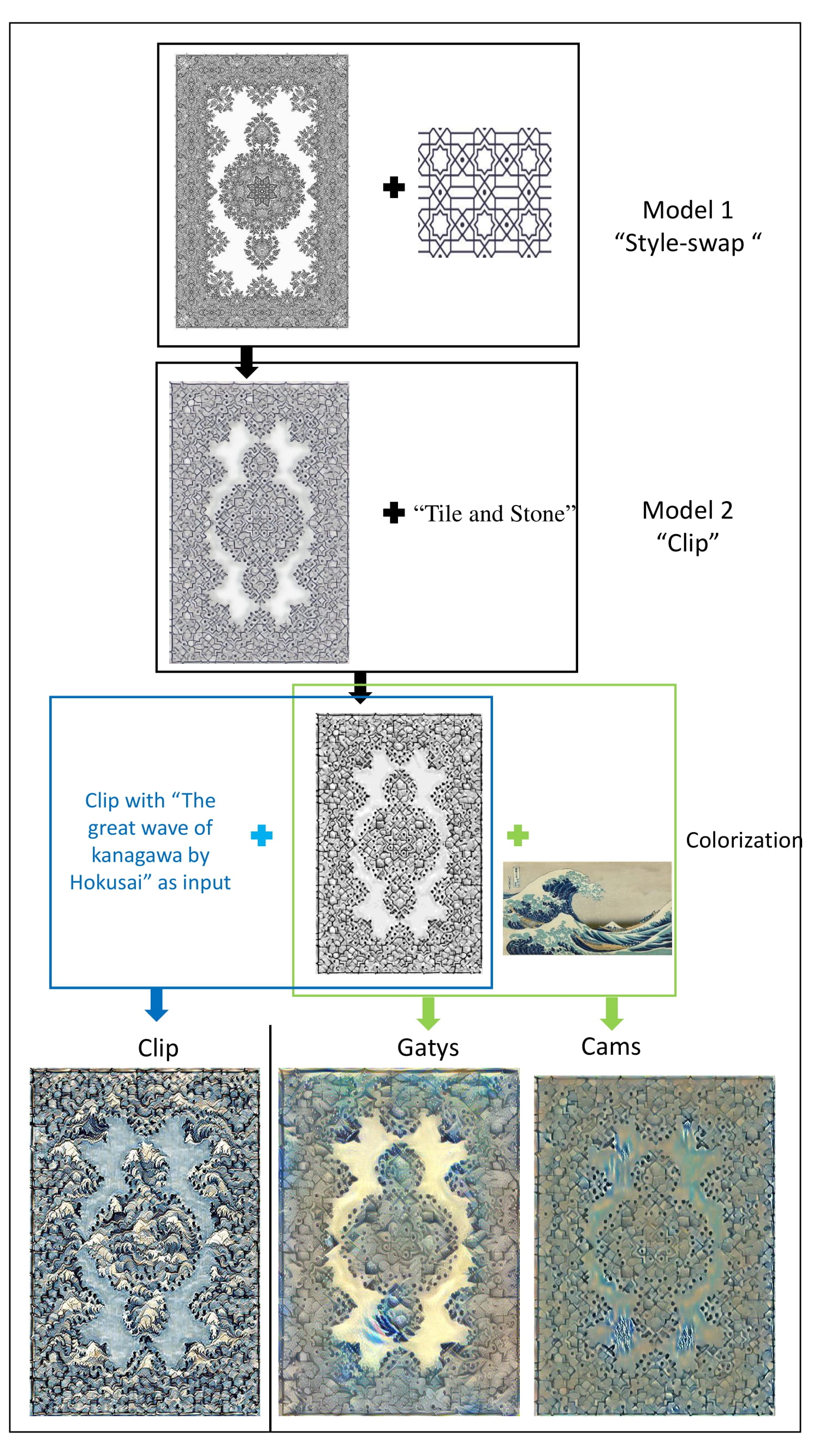}
\includegraphics[width=5cm]{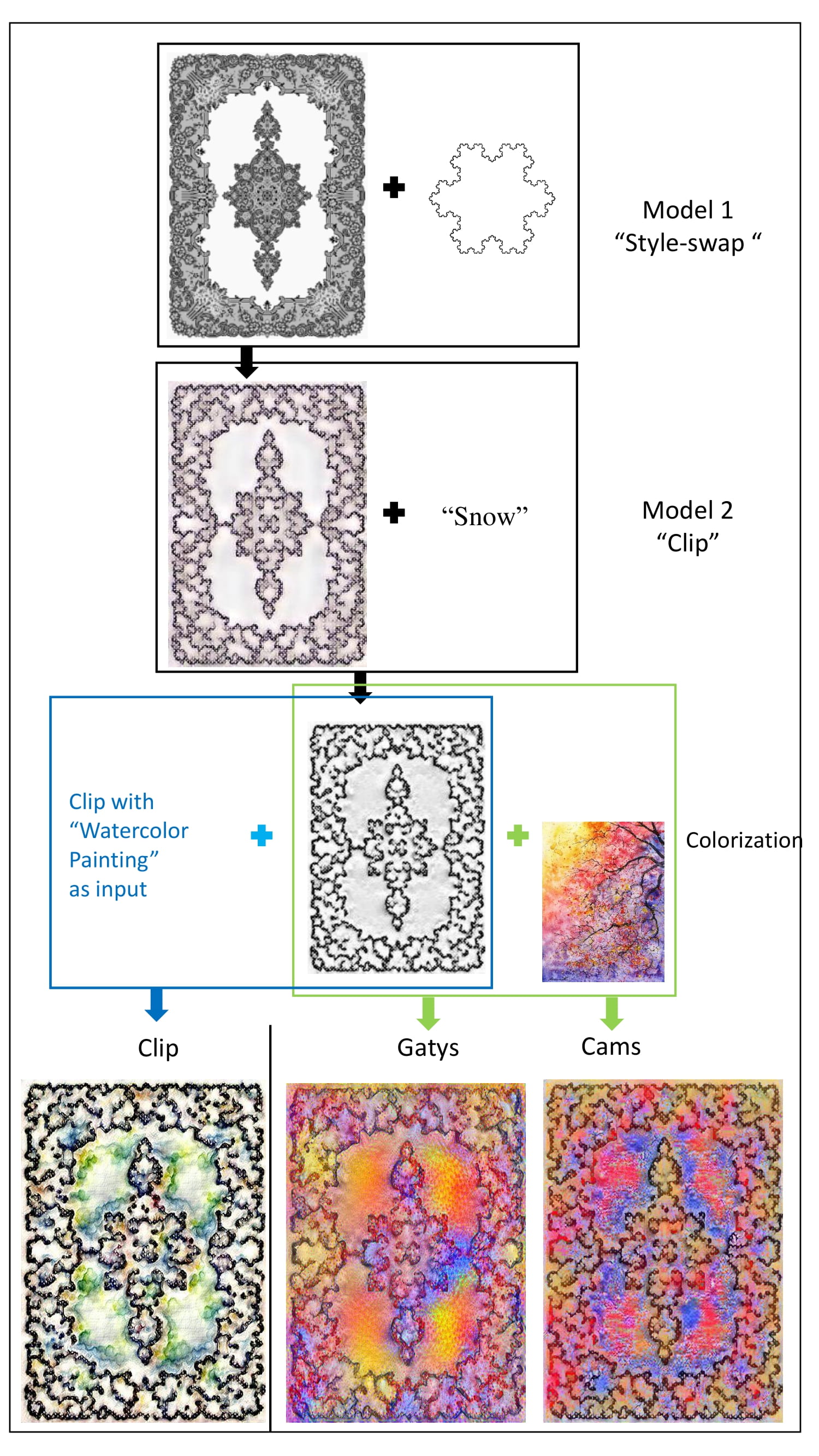}
\caption{Results of generating and coloring new carpet maps when the first method is "Style-Swap" and the second is "Clip-Styler".}
\label{fig:swap-clip}
\end{figure*}

As mentioned in Sec. \ref{sec:generating}, to produce a new carpet map, we employed two style transfer methods consecutively. Due to the order and type of used methods, figures \ref{fig:swap-gatys}, \ref{fig:swap-swap}, and \ref{fig:swap-clip} have displayed the results.
Also, the generated carpet maps, are colorized using three methods mentioned in Sec. \ref{sec:colorization}. to distinguish colorizing methods based on their input style type(image/text), two boxes(green/blue) are used. Furthermore, a better presentation of the final results(color carpet map) is illustrated in Appendix Sec.\ref{app}.


\begin{figure*}[h!]
\captionsetup[subfigure]{labelformat=empty}
\centering
	\subfloat[]{\includegraphics[width=\textwidth/2]{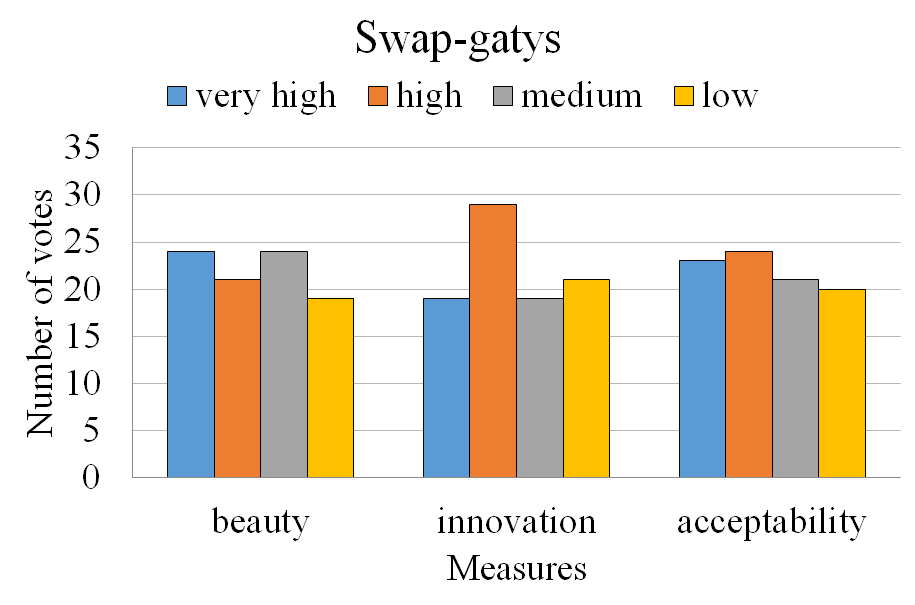}}
    \subfloat[]{\includegraphics[width=\textwidth/2]{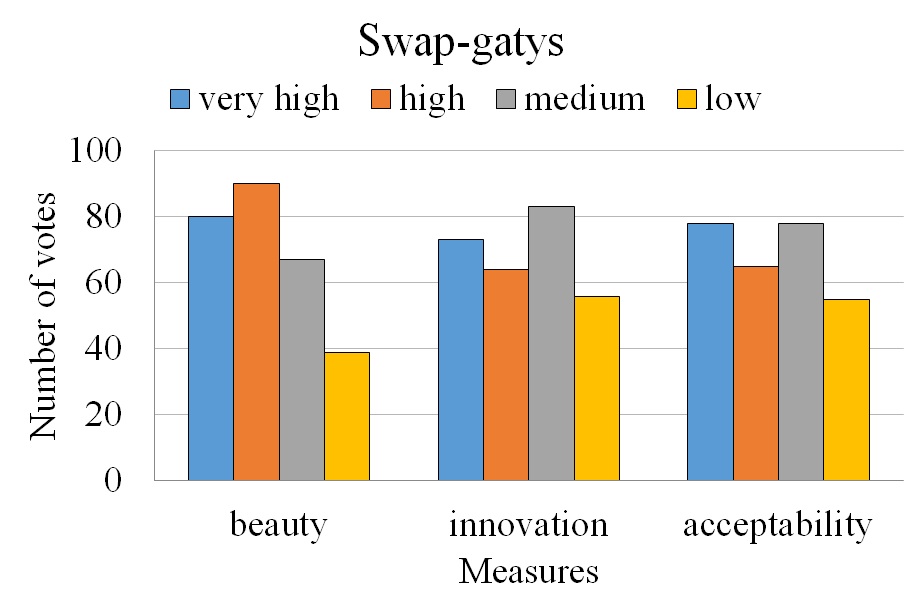}}\\
	\subfloat[]{\includegraphics[width=\textwidth/2]{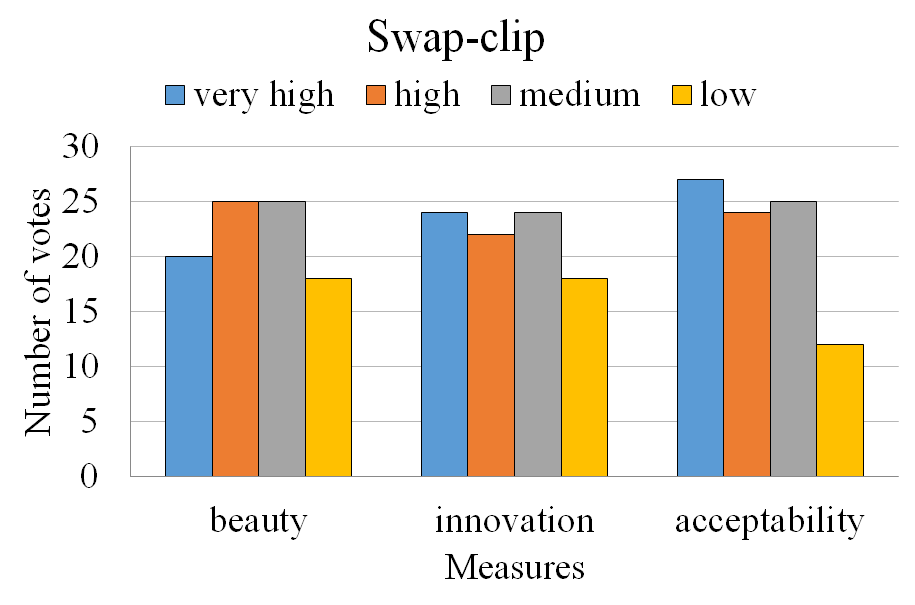}}
    \subfloat[]{\includegraphics[width=\textwidth/2]{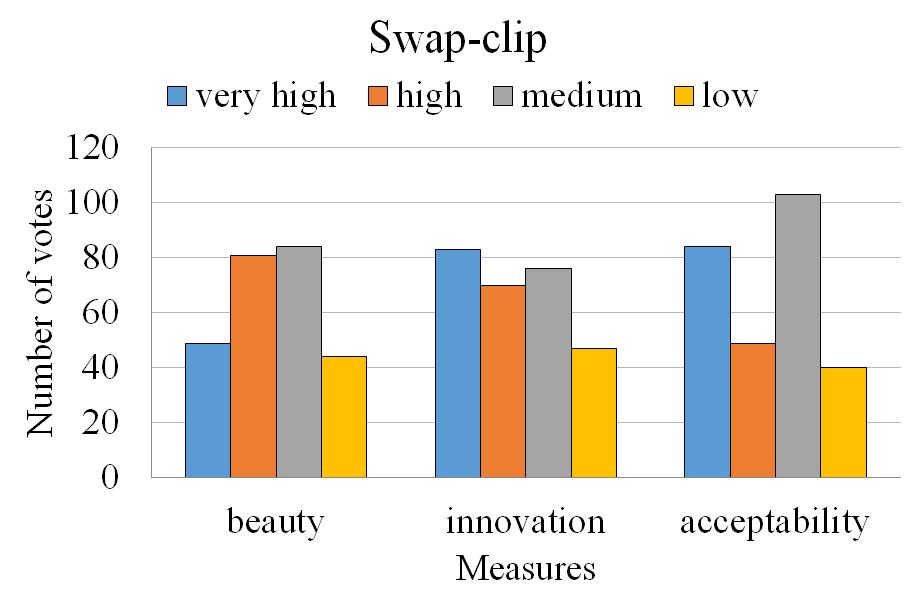}}\\
	\subfloat[]{\includegraphics[width=\textwidth/2]{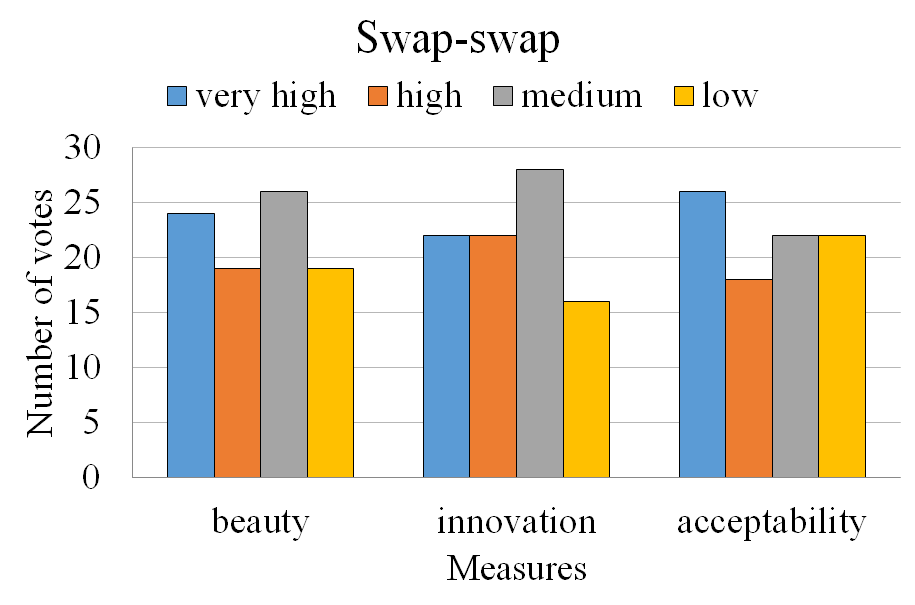}}
	\subfloat[]{\includegraphics[width=\textwidth/2]{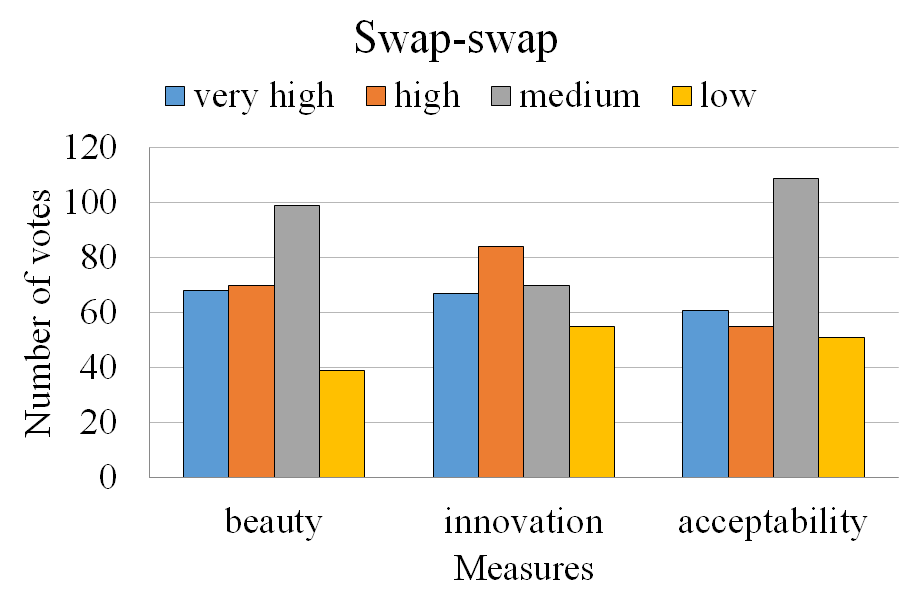}}
\caption{The results of the first user study which is related to the Generating Grayscale Carpet Map stage are illustrated. The first column shows the Experts' votes and the other column demonstrates the General Audience's opinion. In this evaluation, 28 carpet experts and 138 general audiences participated. Also, Measures point to the three concepts(beauty, innovation, and acceptability) that user study questions were about them.}
\label{fig:firstform}
\end{figure*}

\begin{figure*}[h!]
    \captionsetup[subfigure]{labelformat=empty}
	\centering
    \subfloat[]{\includegraphics[width=\textwidth]{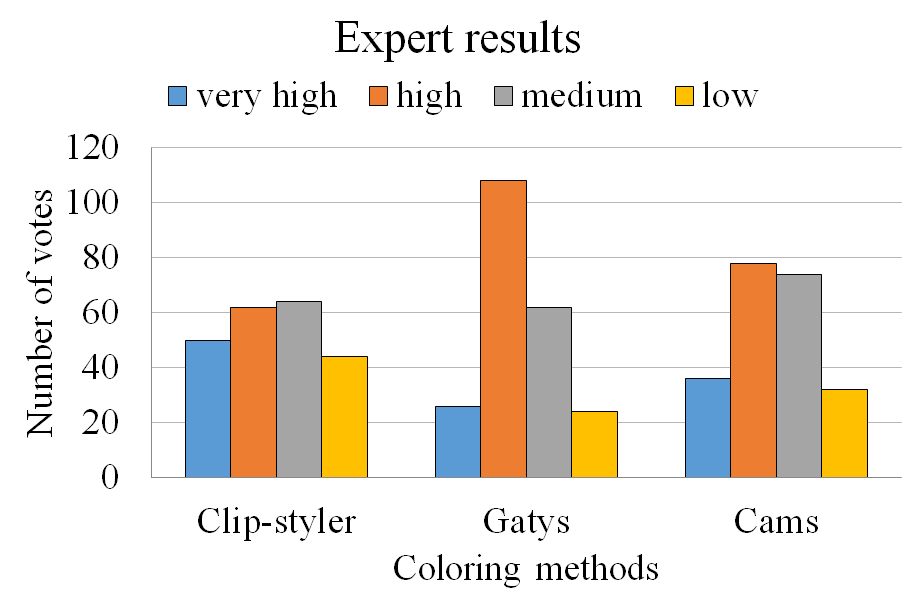}} \\
    \subfloat[]{\includegraphics[width=\textwidth]{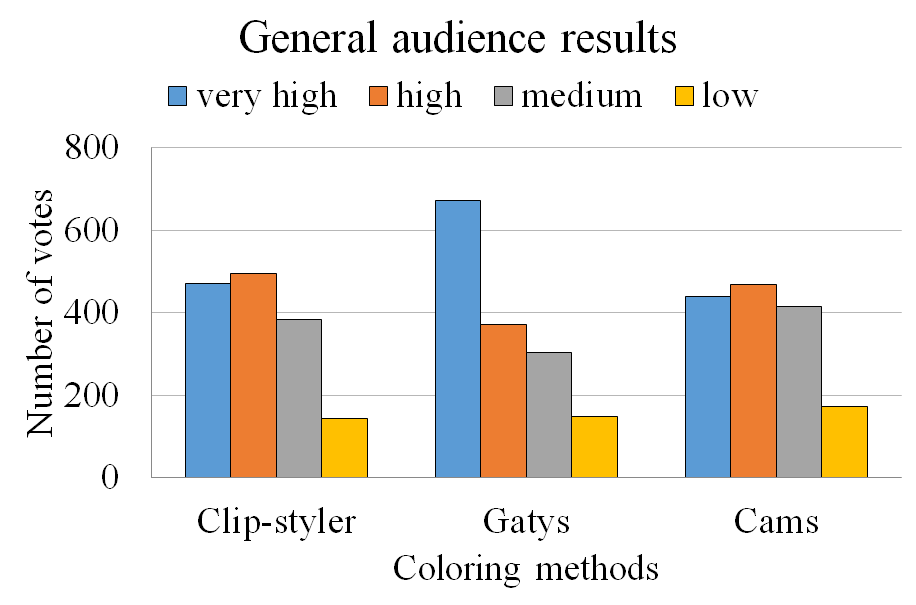}} \\
	\caption{The results of the second user study which is related to the Colorizing Carpet Map phase are given. in which 20 carpet experts and 136 general audiences participate. There were 11 questions and four levels(Very High, High, Medium, and low) as answers for each coloring method. The number of votes is the total votes of each level (Very High, High, Medium, and low) for each method.}
	\label{fig:secondform}
\end{figure*}

To evaluate the proposed methods, user studies have been used. In this evaluation, two groups, the general audience and carpet experts participated in the user studies.
Moreover, two evaluation forms have been conducted, one for the new grayscale designs produced by the three proposed methods and the other for the colored designs with the three methods mentioned in Sec. \ref{sec:colorization}. The first user study form is about three concepts which are the level of beauty, innovation, and acceptability. Also, This form includes 6 questions per proposed method.
In figure \ref{fig:firstform} the results obtained from this user study are shown. As mentioned above, the second user study asks the users about the beauty and innovation of colorizing the new carpet maps. This form contains 11 questions about each of the three proposed methods. Fig. \ref{fig:secondform} has illustrated the results of this form. In both user-study forms were considered four levels (Very High, High, Medium, low ) as answers to the questions. Moreover,  We briefly name our proposed methods for better display. Swap-Gatys means that the Style-Swap and Gatys methods are applied consecutively, Swap-Swap means the Style-Swap method is applied twice, and Swap-Clip means the Style-Swap and Clip-Styler methods are employed serially.

According to the results of the first user study which are related to the step of creating new grayscale carpet maps (figure\ref{fig:firstform}), matching with expert's and general audience's viewpoints, the carpet maps produced by the Swap-Gates method have had more votes in the beauty concept. Also, the two groups believed that the designs of the Swap-Clip method have been more innovative. Their opinion about acceptability is opposite to each other. Due to the experts' votes, the maps of the Swap-Swap method were more acceptable. However, matching the opinion of the general audience, the maps of the Swap-Gates method have been more acceptable.
The results of the second user study(figure\ref{fig:secondform}) which is about the colorizing stage, show two groups (experts and general audience ) almost had the same opinion about the three colorization methods. According to their viewpoint, Gatys, Clip-Styler, and Cams had better performance in coloring respectively.
To recap the results of the user study, the Swap-Gatys method makes the most beautiful grayscale carpet design, the Swap-Clip generates the most innovative maps, and finally, the Gatys method has better performance in coloring. Therefore, our proposed methods on two-step (generating new carpet maps and colorization them) have achieved favorable opinions from the congregation.

\clearpage

\section{Conclusion}
In this paper, we proposed some methods to create modern carpet maps based on neural style transfer methods. Moreover, some style transfer methods have been introduced for colorizing new carpet maps by trial and error. As mentioned earlier, to create more different carpet designs two style transfer methods were employed consecutively. In addition, to generate various maps three different methods were used as the second method separately. To recap, the most significant outcome of the proposed methods was generating new and various carpet maps much faster than traditional and manual ways without much human intervention. Furthermore, according to user study results that have been obtained from general audiences and carpet experts, our generated carpet maps have achieved desirable outcomes from beauty, acceptability, and innovation aspects. In addition, in this research work specific patterns and style transfer methods were employed while there have been various patterns and new style transfer models that can be utilized for future works in this field.

\section{acknowledgment}
This research has been supported by the Iran National Science
Foundation (INSF) and Shahid Bahonar University of Kerman (grant number: 99029504).

\bibliographystyle{IEEEtran}
\bibliography{ref}

\begin{thebibliography}{10}
\providecommand{\url}[1]{#1}
\csname url@samestyle\endcsname
\providecommand{\newblock}{\relax}
\providecommand{\bibinfo}[2]{#2}
\providecommand{\BIBentrySTDinterwordspacing}{\spaceskip=0pt\relax}
\providecommand{\BIBentryALTinterwordstretchfactor}{4}
\providecommand{\BIBentryALTinterwordspacing}{\spaceskip=\fontdimen2\font plus
\BIBentryALTinterwordstretchfactor\fontdimen3\font minus
  \fontdimen4\font\relax}
\providecommand{\BIBforeignlanguage}[2]{{%
\expandafter\ifx\csname l@#1\endcsname\relax
\typeout{** WARNING: IEEEtran.bst: No hyphenation pattern has been}%
\typeout{** loaded for the language `#1'. Using the pattern for}%
\typeout{** the default language instead.}%
\else
\language=\csname l@#1\endcsname
\fi
#2}}
\providecommand{\BIBdecl}{\relax}
\BIBdecl

\bibitem{gatys2016image}
L.~A. Gatys, A.~S. Ecker, and M.~Bethge, ``Image style transfer using
  convolutional neural networks,'' in \emph{Proceedings of the IEEE conference
  on computer vision and pattern recognition}, 2016, pp. 2414--2423.

\bibitem{ashikhmin2003fast}
N.~Ashikhmin, ``Fast texture transfer,'' \emph{IEEE computer Graphics and
  Applications}, vol.~23, no.~4, pp. 38--43, 2003.

\bibitem{zhao2020survey}
C.~Zhao, ``A survey on image style transfer approaches using deep learning,''
  in \emph{Journal of Physics: Conference Series}, vol. 1453, no.~1.\hskip 1em
  plus 0.5em minus 0.4em\relax IOP Publishing, 2020, p. 012129.

\bibitem{sheng2018avatar}
L.~Sheng, Z.~Lin, J.~Shao, and X.~Wang, ``Avatar-net: Multi-scale zero-shot
  style transfer by feature decoration,'' in \emph{Proceedings of the IEEE
  conference on computer vision and pattern recognition}, 2018, pp. 8242--8250.

\bibitem{gu2018arbitrary}
S.~Gu, C.~Chen, J.~Liao, and L.~Yuan, ``Arbitrary style transfer with deep
  feature reshuffle,'' in \emph{Proceedings of the IEEE Conference on Computer
  Vision and Pattern Recognition}, 2018, pp. 8222--8231.

\bibitem{li2016combining}
C.~Li and M.~Wand, ``Combining markov random fields and convolutional neural
  networks for image synthesis,'' in \emph{Proceedings of the IEEE conference
  on computer vision and pattern recognition}, 2016, pp. 2479--2486.

\bibitem{li2017universal}
Y.~Li, C.~Fang, J.~Yang, Z.~Wang, X.~Lu, and M.-H. Yang, ``Universal style
  transfer via feature transforms,'' \emph{Advances in neural information
  processing systems}, vol.~30, 2017.

\bibitem{li2019learning}
X.~Li, S.~Liu, J.~Kautz, and M.-H. Yang, ``Learning linear transformations for
  fast image and video style transfer,'' in \emph{Proceedings of the IEEE/CVF
  Conference on Computer Vision and Pattern Recognition}, 2019, pp. 3809--3817.

\bibitem{huang2017arbitrary}
X.~Huang and S.~Belongie, ``Arbitrary style transfer in real-time with adaptive
  instance normalization,'' in \emph{Proceedings of the IEEE international
  conference on computer vision}, 2017, pp. 1501--1510.

\bibitem{zhang2022domain}
Y.~Zhang, F.~Tang, W.~Dong, H.~Huang, C.~Ma, T.-Y. Lee, and C.~Xu, ``Domain
  enhanced arbitrary image style transfer via contrastive learning,''
  \emph{arXiv preprint arXiv:2205.09542}, 2022.

\bibitem{chen2016fast}
T.~Q. Chen and M.~Schmidt, ``Fast patch-based style transfer of arbitrary
  style,'' \emph{arXiv preprint arXiv:1612.04337}, 2016.

\bibitem{zhang2019multimodal}
Y.~Zhang, C.~Fang, Y.~Wang, Z.~Wang, Z.~Lin, Y.~Fu, and J.~Yang, ``Multimodal
  style transfer via graph cuts,'' in \emph{Proceedings of the IEEE/CVF
  International Conference on Computer Vision}, 2019, pp. 5943--5951.

\bibitem{afifi2021cams}
M.~Afifi, A.~Abuolaim, M.~Hussien, M.~A. Brubaker, and M.~S. Brown, ``Cams:
  Color-aware multi-style transfer,'' \emph{arXiv preprint arXiv:2106.13920},
  2021.

\bibitem{kwon2022clipstyler}
G.~Kwon and J.~C. Ye, ``Clipstyler: Image style transfer with a single text
  condition,'' in \emph{Proceedings of the IEEE/CVF Conference on Computer
  Vision and Pattern Recognition}, 2022, pp. 18\,062--18\,071.

\bibitem{liu2022name}
Z.-S. Liu, L.-W. Wang, W.-C. Siu, and V.~Kalogeiton, ``Name your style: An
  arbitrary artist-aware image style transfer,'' \emph{arXiv preprint
  arXiv:2202.13562}, 2022.

\bibitem{radford2021learning}
A.~Radford, J.~W. Kim, C.~Hallacy, A.~Ramesh, G.~Goh, S.~Agarwal, G.~Sastry,
  A.~Askell, P.~Mishkin, J.~Clark \emph{et~al.}, ``Learning transferable visual
  models from natural language supervision,'' in \emph{International Conference
  on Machine Learning}.\hskip 1em plus 0.5em minus 0.4em\relax PMLR, 2021, pp.
  8748--8763.

\bibitem{karras2019style}
T.~Karras, S.~Laine, and T.~Aila, ``A style-based generator architecture for
  generative adversarial networks,'' in \emph{Proceedings of the IEEE/CVF
  conference on computer vision and pattern recognition}, 2019, pp. 4401--4410.

\bibitem{zhu2017unpaired}
J.-Y. Zhu, T.~Park, P.~Isola, and A.~A. Efros, ``Unpaired image-to-image
  translation using cycle-consistent adversarial networks,'' in
  \emph{Proceedings of the IEEE international conference on computer vision},
  2017, pp. 2223--2232.

\bibitem{chang2015palette}
H.~Chang, O.~Fried, Y.~Liu, S.~DiVerdi, and A.~Finkelstein, ``Palette-based
  photo recoloring.'' \emph{ACM Trans. Graph.}, vol.~34, no.~4, pp. 139--1,
  2015.

\bibitem{gal2021stylegan}
R.~Gal, O.~Patashnik, H.~Maron, G.~Chechik, and D.~Cohen-Or, ``Stylegan-nada:
  Clip-guided domain adaptation of image generators,'' \emph{arXiv preprint
  arXiv:2108.00946}, 2021.

\bibitem{lin2014microsoft}
T.-Y. Lin, M.~Maire, S.~Belongie, J.~Hays, P.~Perona, D.~Ramanan,
  P.~Doll{\'a}r, and C.~L. Zitnick, ``Microsoft coco: Common objects in
  context,'' in \emph{European conference on computer vision}.\hskip 1em plus
  0.5em minus 0.4em\relax Springer, 2014, pp. 740--755.

\bibitem{krishna2017visual}
R.~Krishna, Y.~Zhu, O.~Groth, J.~Johnson, K.~Hata, J.~Kravitz, S.~Chen,
  Y.~Kalantidis, L.-J. Li, D.~A. Shamma \emph{et~al.}, ``Visual genome:
  Connecting language and vision using crowdsourced dense image annotations,''
  \emph{International journal of computer vision}, vol. 123, no.~1, pp. 32--73,
  2017.

\bibitem{thomee2016yfcc100m}
B.~Thomee, D.~A. Shamma, G.~Friedland, B.~Elizalde, K.~Ni, D.~Poland, D.~Borth,
  and L.-J. Li, ``Yfcc100m: The new data in multimedia research,''
  \emph{Communications of the ACM}, vol.~59, no.~2, pp. 64--73, 2016.

\end{thebibliography}
\newpage
\section{Appendix}\label{app}
In the following, a better display of the final results is given in the Sec. \ref{exp}. 

\begin{figure*}[h!]
	\centering
	\resizebox{0.3\linewidth}{.27\textheight}{\includegraphics[width=\textwidth]{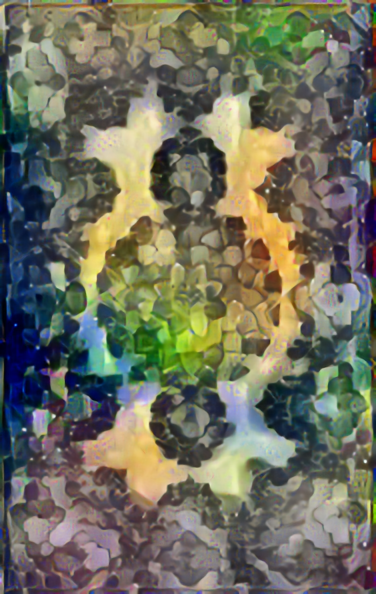}}
	\resizebox{0.3\linewidth}{.27\textheight}{\includegraphics[width=\textwidth]{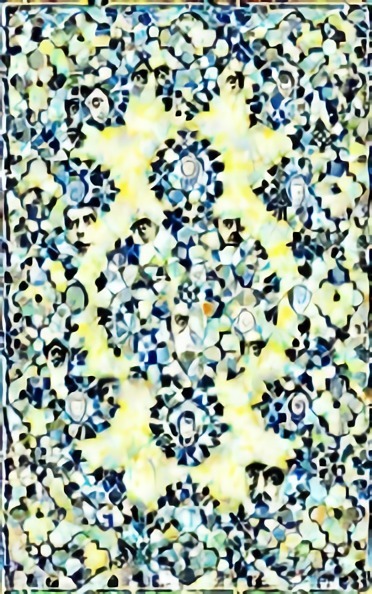}}
	\resizebox{0.3\linewidth}{.27\textheight}{\includegraphics[width=\textwidth]{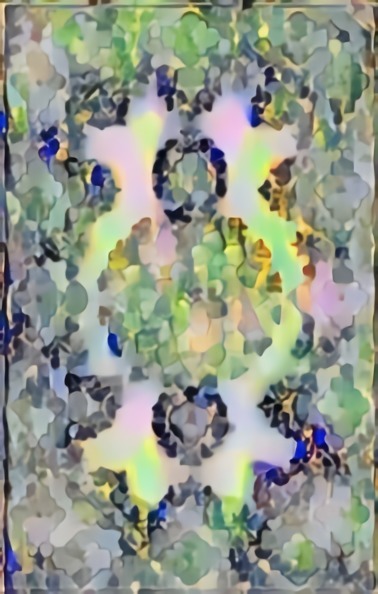}}\\
	\resizebox{0.3\linewidth}{.27\textheight}{\includegraphics[width=\textwidth]{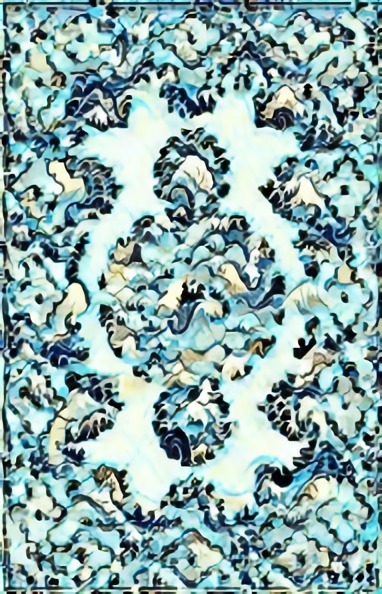}}
	\resizebox{0.3\linewidth}{.27\textheight}{\includegraphics[width=\textwidth]{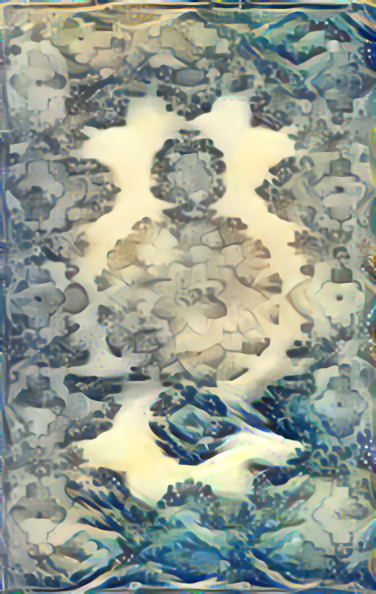}}
	\resizebox{0.3\linewidth}{.27\textheight}{\includegraphics[width=\textwidth]{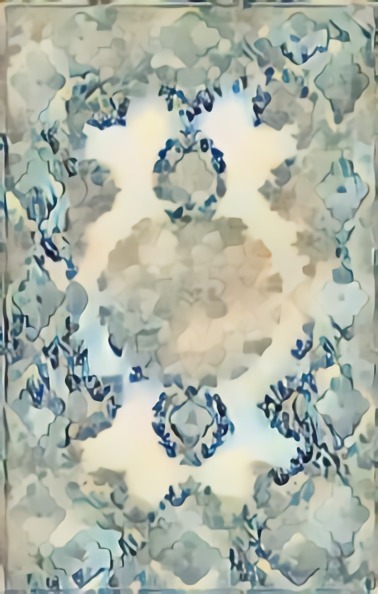}}\\
	\resizebox{0.3\linewidth}{.27\textheight}{\includegraphics[width=\textwidth]{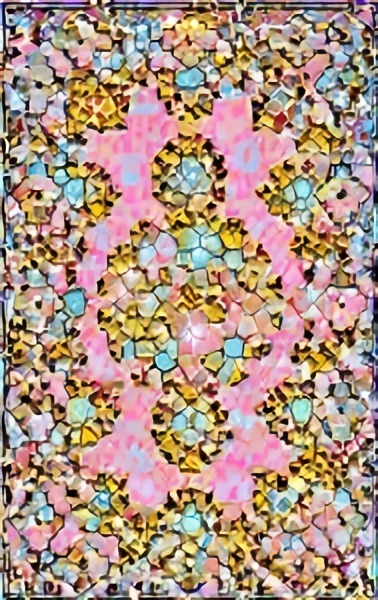}}
	\resizebox{0.3\linewidth}{.27\textheight}{\includegraphics[width=\textwidth]{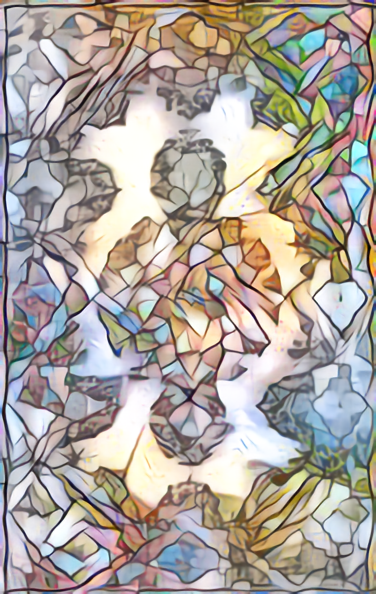}}
	\resizebox{0.3\linewidth}{.27\textheight}{\includegraphics[width=\textwidth]{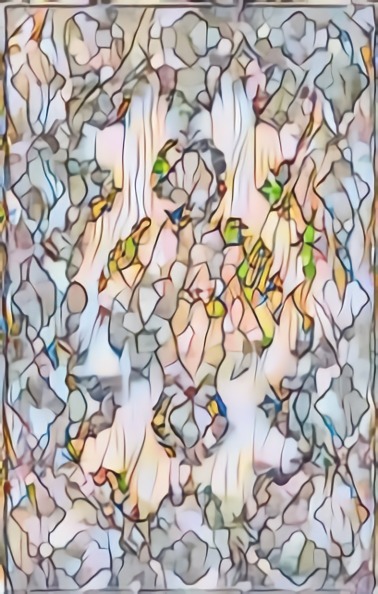}}\\
	\resizebox{0.3\linewidth}{.27\textheight}{\includegraphics[width=\textwidth]{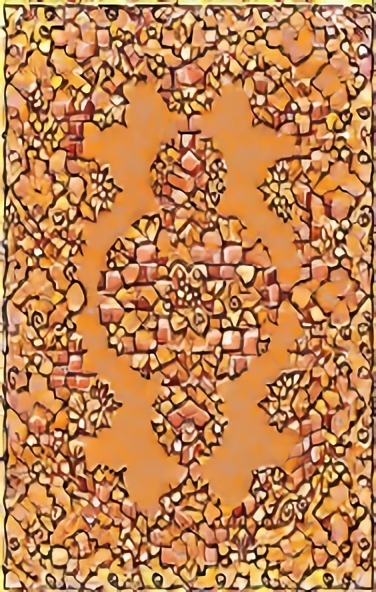}}
	\resizebox{0.3\linewidth}{.27\textheight}{\includegraphics[width=\textwidth]{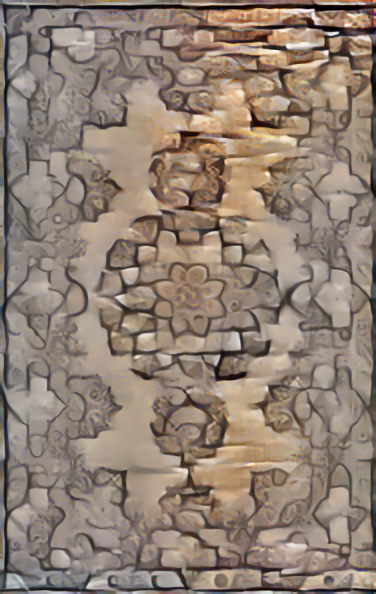}}
	\resizebox{0.3\linewidth}{.27\textheight}{\includegraphics[width=\textwidth]{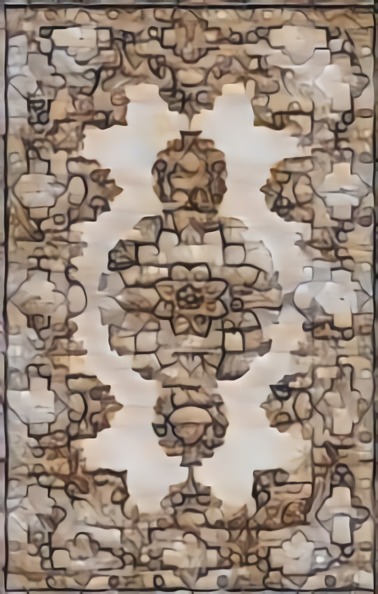}}
	\caption{The final results of the Swap-Gatys method}
	\label{app1}
\end{figure*}

\begin{figure*}[h!]
	\centering
	\resizebox{0.3\linewidth}{.27\textheight}{\includegraphics[width=\textwidth]{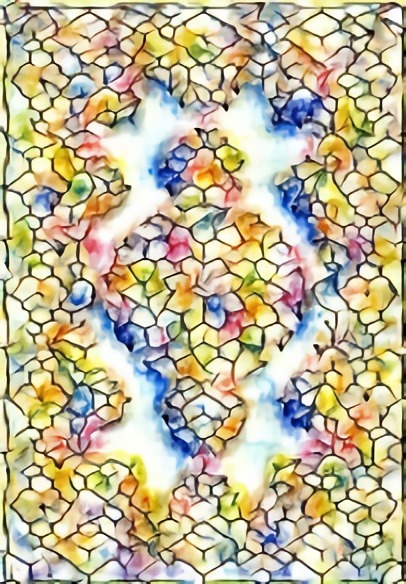}}
	\resizebox{0.3\linewidth}{.27\textheight}{\includegraphics[width=\textwidth]{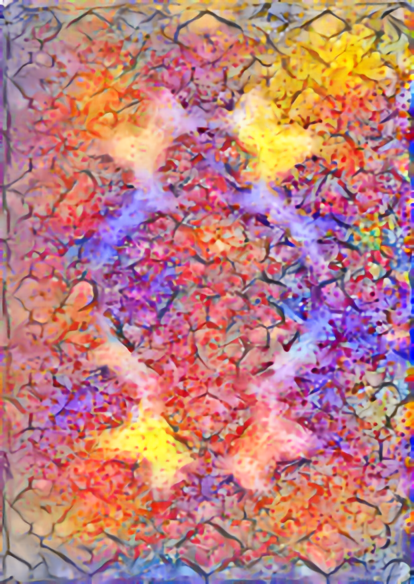}}
	\resizebox{0.3\linewidth}{.27\textheight}{\includegraphics[width=\textwidth]{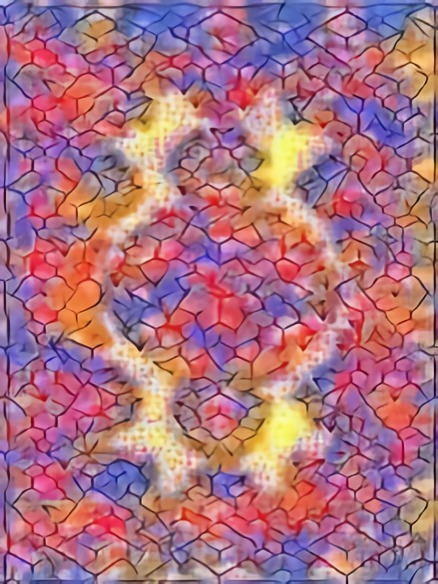}}\\
	\resizebox{0.3\linewidth}{.27\textheight}{\includegraphics[width=\textwidth]{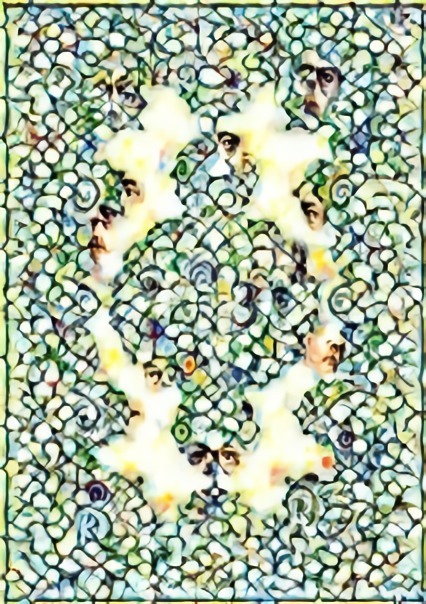}}
	\resizebox{0.3\linewidth}{.27\textheight}{\includegraphics[width=\textwidth]{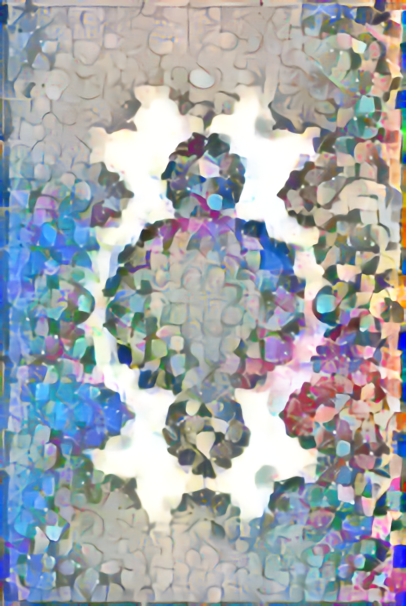}}
	\resizebox{0.3\linewidth}{.27\textheight}{\includegraphics[width=\textwidth]{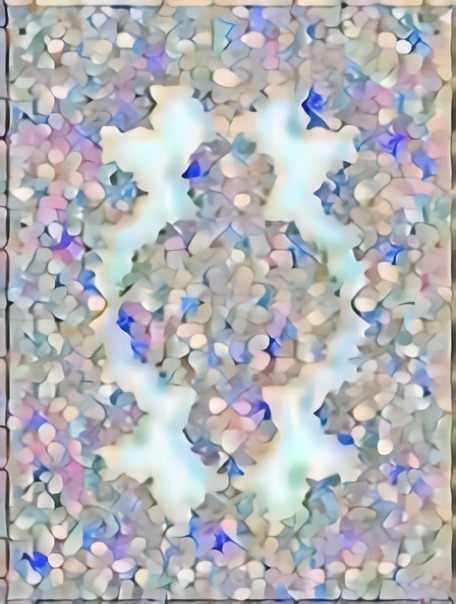}}\\
	\resizebox{0.3\linewidth}{.27\textheight}{\includegraphics[width=\textwidth]{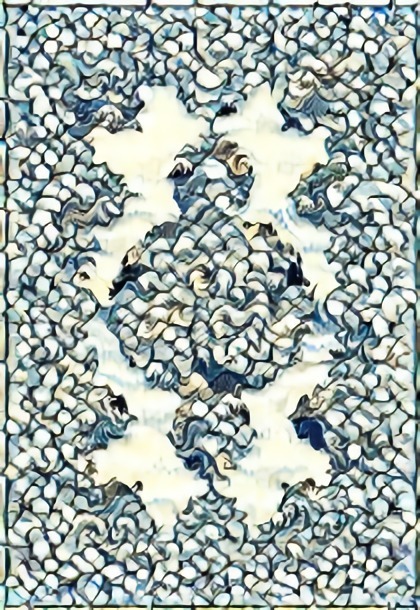}}
	\resizebox{0.3\linewidth}{.27\textheight}{\includegraphics[width=\textwidth]{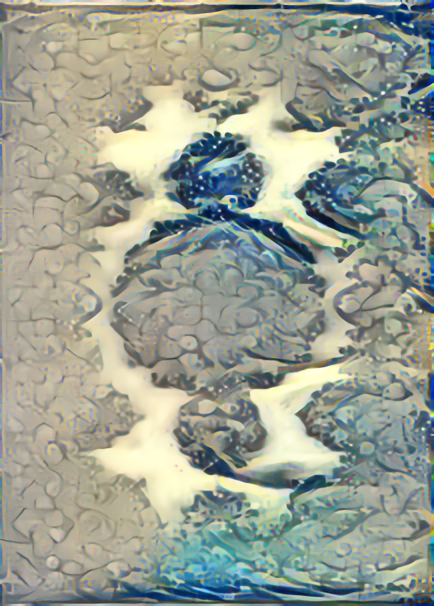}}
	\resizebox{0.3\linewidth}{.27\textheight}{\includegraphics[width=\textwidth]{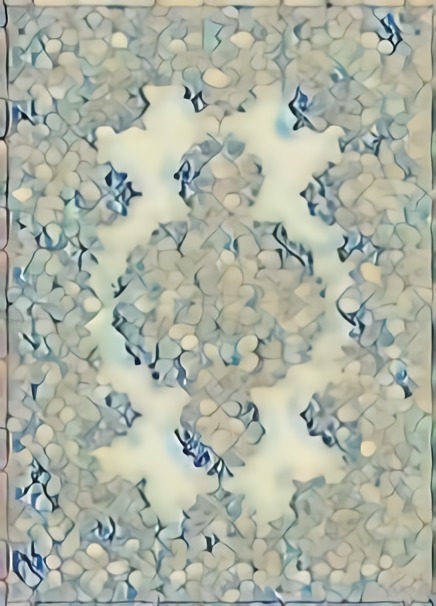}}\\
	\resizebox{0.3\linewidth}{.27\textheight}{\includegraphics[width=\textwidth]{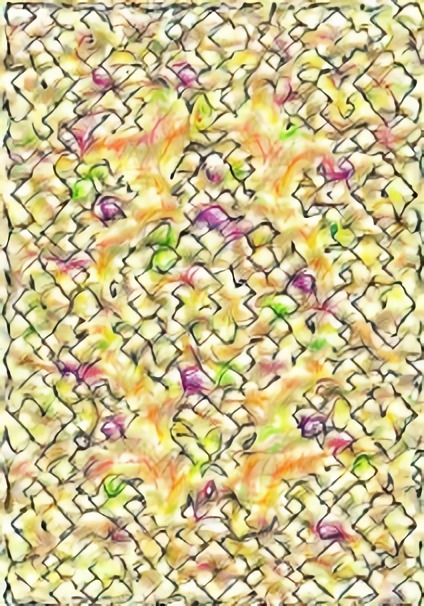}}
	\resizebox{0.3\linewidth}{.27\textheight}{\includegraphics[width=\textwidth]{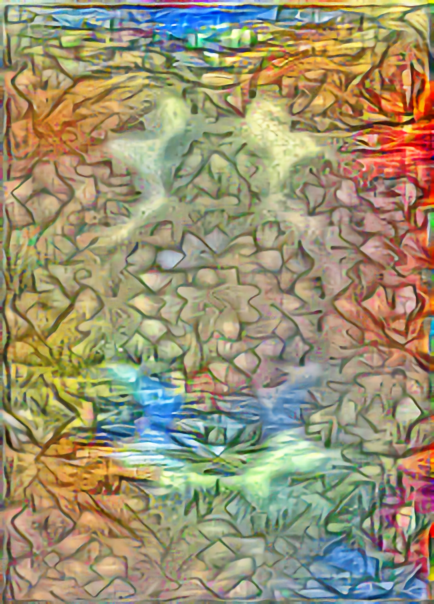}}
	\resizebox{0.3\linewidth}{.27\textheight}{\includegraphics[width=\textwidth]{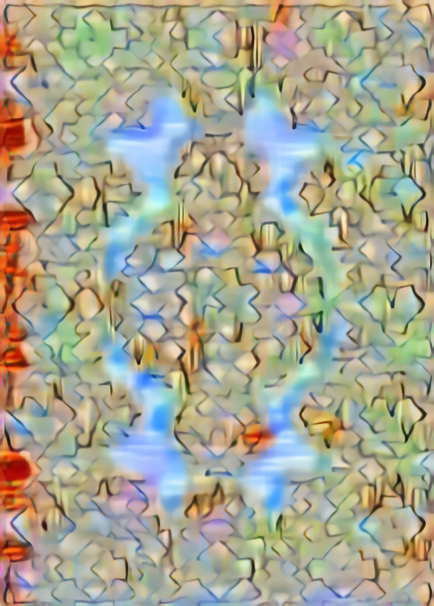}}
	\caption{The final results of the Swap-Swap method}
	\label{app2}
\end{figure*}

\begin{figure*}[h!]
	\centering
	\resizebox{0.3\linewidth}{.27\textheight}{\includegraphics[width=\textwidth]{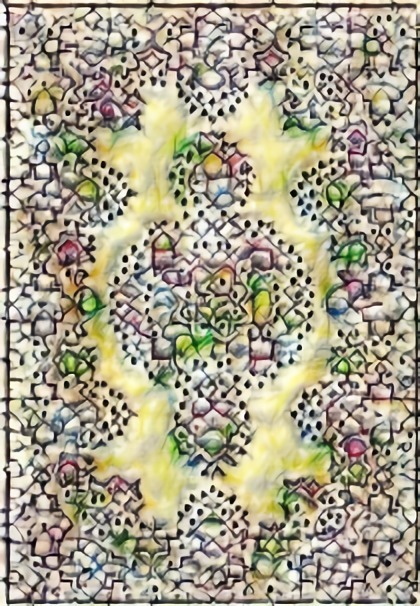}}
	\resizebox{0.3\linewidth}{.27\textheight}{\includegraphics[width=\textwidth]{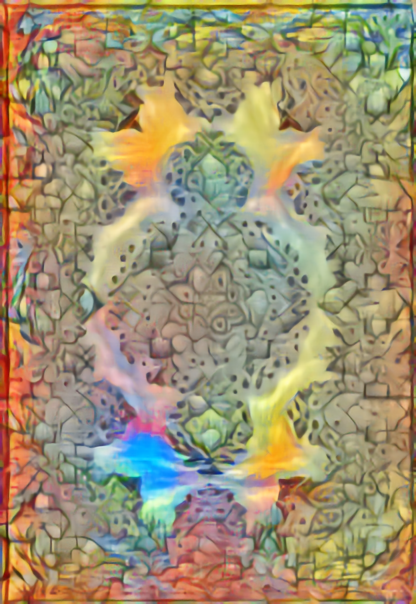}}
	\resizebox{0.3\linewidth}{.27\textheight}{\includegraphics[width=\textwidth]{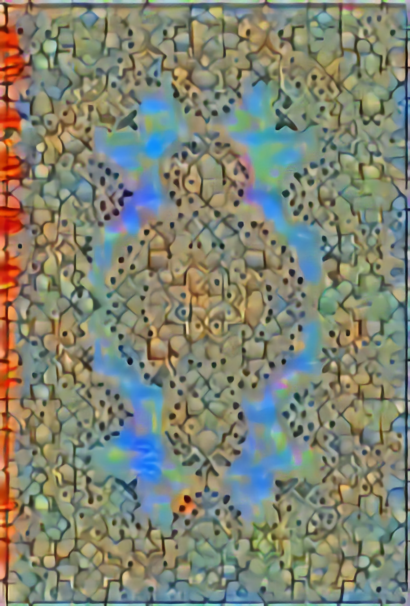}}\\
	\resizebox{0.3\linewidth}{.27\textheight}{\includegraphics[width=\textwidth]{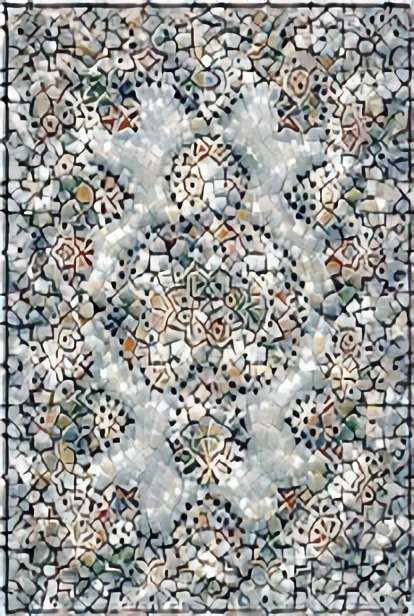}}
	\resizebox{0.3\linewidth}{.27\textheight}{\includegraphics[width=\textwidth]{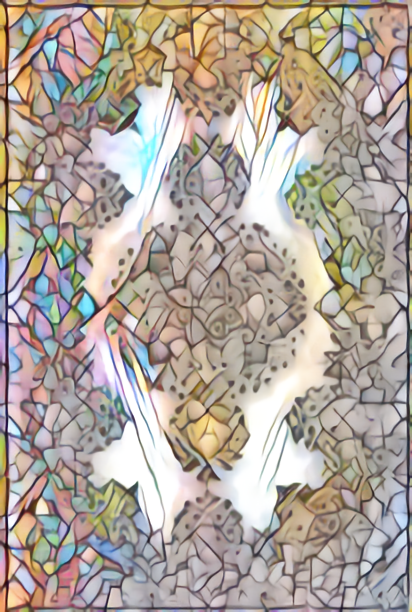}}
	\resizebox{0.3\linewidth}{.27\textheight}{\includegraphics[width=\textwidth]{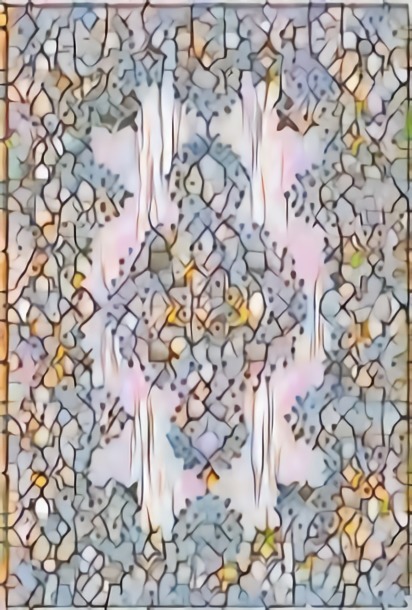}}\\
	\resizebox{0.3\linewidth}{.27\textheight}{\includegraphics[width=\textwidth]{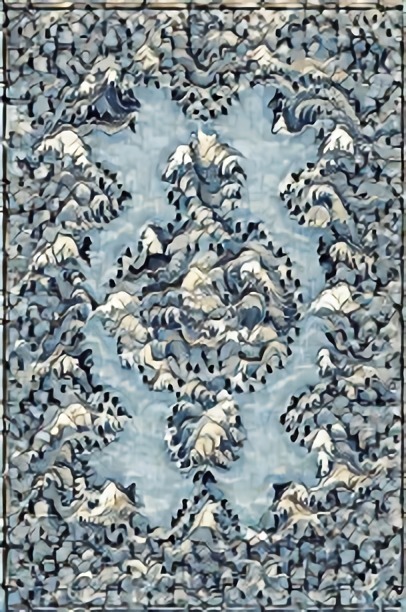}}
	\resizebox{0.3\linewidth}{.27\textheight}{\includegraphics[width=\textwidth]{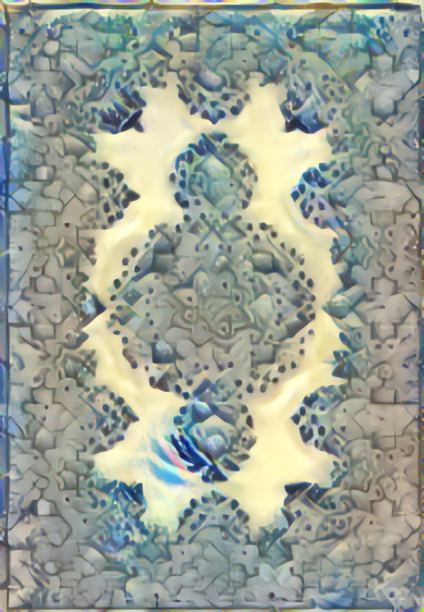}}
	\resizebox{0.3\linewidth}{.27\textheight}{\includegraphics[width=\textwidth]{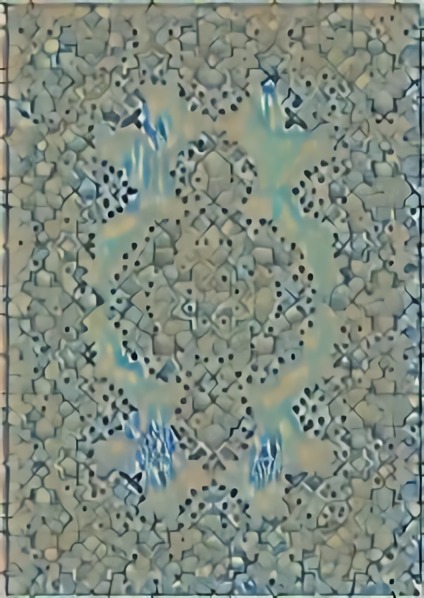}}\\
	\resizebox{0.3\linewidth}{.27\textheight}{\includegraphics[width=\textwidth]{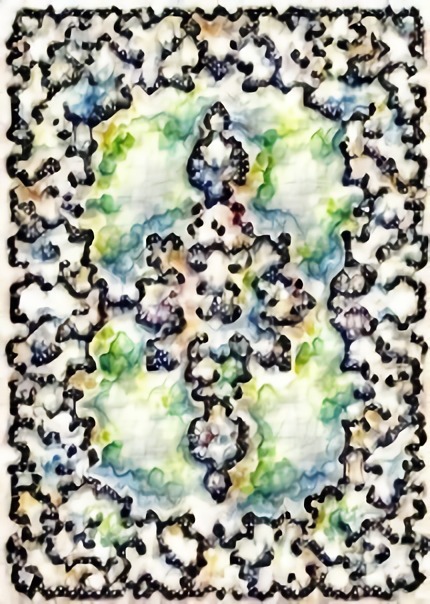}}
	\resizebox{0.3\linewidth}{.27\textheight}{\includegraphics[width=\textwidth]{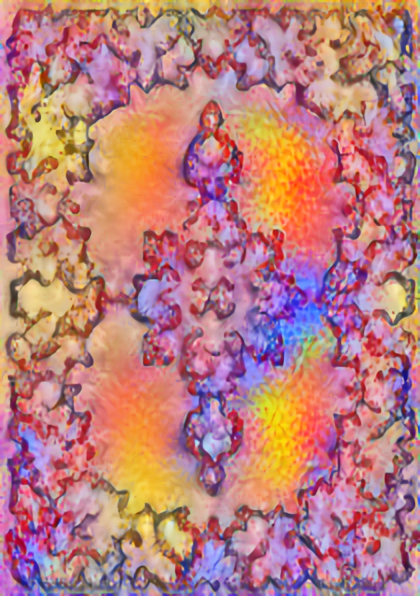}}
	\resizebox{0.3\linewidth}{.27\textheight}{\includegraphics[width=\textwidth]{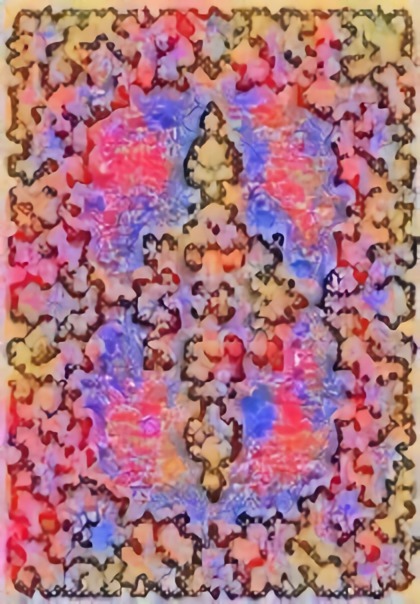}}
	\caption{The final results of the Swap-Clip method}
	\label{app3}
\end{figure*}

\clearpage
\end{document}